\documentclass[times,twocolumn,final]{elsarticle}

\usepackage{medima}
\usepackage{framed,multirow}

\usepackage{amssymb}
\usepackage{latexsym}

\usepackage[colorlinks=true]{hyperref}
\usepackage{hyperref}
\usepackage{graphicx}
\usepackage{amsmath,amssymb} 
\usepackage{color}
\usepackage{xcolor}
\usepackage[font=small,labelfont=bf]{caption}

\usepackage{makeidx}  
\usepackage{color}
\usepackage{graphicx}

\usepackage{amsmath}
\usepackage{amssymb}
\usepackage[ruled,lined,linesnumbered]{algorithm2e}
\usepackage{algorithmic}
\usepackage{blindtext}
\usepackage{enumitem}
\usepackage{multirow}
\usepackage[labelformat=simple]{subfig}
\usepackage{threeparttable/threeparttable}
\usepackage{bm}
\usepackage[normalem]{ulem}
\usepackage{hyperref}
\usepackage{multirow, makecell}

\usepackage{colortbl}
\definecolor{maroon}{rgb}{0.816, 0.208, 0.188}
\definecolor{iblue}{rgb}{0.06, 0.75, 1.0}

\usepackage{amssymb}
\usepackage{pifont}
\newcommand{\xmark}{\ding{55}}%

\newcommand{\ie}{\mbox{\emph{i.e.,\ }}}

\usepackage{array}
\newcolumntype{P}[1]{>{\centering\arraybackslash}p{#1}}

\usepackage{soul}

\newcommand{\eg}{\mbox{\emph{e.g.,\ }}}

\newlength\savewidth\newcommand\shline{\noalign{\global\savewidth\arrayrulewidth
  \global\arrayrulewidth 0.8pt}\hline\noalign{\global\arrayrulewidth\savewidth}}

\definecolor{newcolor}{rgb}{.8,.349,.1}

\journal{Medical Image Analysis}

\begin{document}

\verso{Zongwei Zhou \textit{et~al.}}

\begin{frontmatter}

\title{Models Genesis}

\author[1]{Zongwei \snm{Zhou}}
\author[2]{Vatsal \snm{Sodha}}
\author[2]{Jiaxuan \snm{Pang}}
\author[3]{Michael B. \snm{Gotway}}
\author[1]{Jianming \snm{Liang}\corref{cor1}}
\cortext[cor1]{Corresponding author: \href{mailto:Jianming.Liang@asu.edu}{Jianming.Liang@asu.edu} (Jianming Liang)}

\address[1]{Department of Biomedical Informatics, Arizona State University, Scottsdale, AZ 85259, USA}
\address[2]{School of Computing, Informatics, and Decision Systems Engineering, Arizona State University, Tempe, AZ 85281 USA}
\address[3]{Department of Radiology, Mayo Clinic, Scottsdale, AZ 85259, USA}

\received{1 May 2019}
\finalform{***}
\accepted{***}
\availableonline{***}
\communicated{***}

\begin{abstract}
Transfer learning from {\em natural} images to {\em medical} images has been established as one of the most practical paradigms in deep learning for medical image analysis. To fit this paradigm, however, 3D imaging tasks in the most prominent imaging modalities (\eg CT and MRI) have to be reformulated and solved in 2D, losing rich 3D anatomical information, thereby inevitably compromising its performance. To overcome this limitation, we have built a set of models, called Generic Autodidactic Models, nicknamed Models Genesis, because they are created {\em ex nihilo} (with no manual labeling), self-taught (learnt by self-supervision), and generic (served as source models for generating application-specific target models). Our extensive experiments demonstrate that our Models Genesis significantly outperform learning from scratch and existing pre-trained 3D models in all five target 3D applications covering both segmentation and classification.
More importantly, learning a model from scratch {\em simply in 3D} may not necessarily yield performance better than transfer learning from ImageNet in 2D, but our Models Genesis consistently top any 2D/2.5D approaches including fine-tuning the models pre-trained from ImageNet as well as fine-tuning the 2D versions of our Models Genesis, confirming the importance of 3D anatomical information and significance of Models Genesis for 3D medical imaging. This performance is attributed to our unified self-supervised learning framework, built on a simple yet powerful observation: the sophisticated and recurrent anatomy in medical images can serve as strong yet free supervision signals for deep models to learn common anatomical representation automatically via self-supervision. 
As open science, all codes and pre-trained Models Genesis are available at \href{https://github.com/MrGiovanni/ModelsGenesis}{https://github.com/MrGiovanni/ModelsGenesis}.
\end{abstract}


\end{frontmatter}


\section{Introduction}
\label{sec:introduction}

\begin{table*}[t]
\centering
\begin{threeparttable}
\footnotesize
\caption{Pre-trained models with proxy tasks and target tasks. This paper uses transfer learning in a broader sense, where a \textit{source model} is first trained to learn image presentation via \textit{full supervision} or \textit{self supervision} by solving a problem, called \textit{proxy task} (general or application-specific), on a \textit{source dataset} with \textit{expert-provided} or \textit{automatically-generated} labels, and then this \textit{pre-trained} source model is fine tuned (transferred) through full supervision to yield a \textit{target model} to solve application-specific problems (\textit{target tasks}) in the same or different datasets (\textit{target datasets}). We refer transfer learning to \textit{same-domain} transfer learning when the models are pre-trained and fine-tuned within the same domain (modality, organ, disease, or dataset), and to \textit{cross-domain} when the models are pre-trained in one domain and fine-tuned for a different domain.
  }
\label{tab:terminology}
    \begin{tabular}{p{0.02\linewidth}p{0.116\linewidth}p{0.055\linewidth}p{0.275\linewidth}p{0.115\linewidth}p{0.275\linewidth}}
        \shline
        \multicolumn{2}{l}{Pre-trained model} & Modality & Source dataset & Superv. / Annot. & Proxy task  \\
        \hline
        \multicolumn{2}{l}{Genesis Chest CT 2D} & CT & LUNA~2016~\citep{setio2017validation} & Self / 0 & Image restoration on 2D Chest CT slices \\
        \multicolumn{2}{l}{Genesis Chest CT (3D)} & CT & LUNA~2016~\citep{setio2017validation} & Self / 0 & Image restoration on 3D Chest CT volumes \\
        \multicolumn{2}{l}{Genesis Chest X-ray (2D)} & X-ray & ChestX-ray8~\citep{wang2017chestx} & Self / 0 & Image restoration on 2D Chest Radiographs \\
        \multicolumn{2}{l}{Models ImageNet} & Natural & ImageNet~\citep{deng2009imagenet} & Full / 14M images & Image classification on 2D ImageNet \\
        \multicolumn{2}{l}{Inflated 3D (I3D)} & Natural & Kinetics~\citep{carreira2017quo} & Full / 240K videos & Action recognition on human action videos \\
        \multicolumn{2}{l}{NiftyNet} & CT & Pancreas-CT \& BTCV~\citep{gibson2018automatic} & Full / 90 cases & Organ segmentation on abdominal CT \\
        \multicolumn{2}{l}{MedicalNet} & CT, MRI & 3DSeg-8~\citep{chen2019med3d} & Full / 1,638 cases & Disease/organ segmentation on 8 datasets \\
        \hline
        \hline
        Code$^{\dagger}$ & Object & Modality & \multicolumn{2}{l}{Target dataset}  & Target task \\
        \hline
        \texttt{NCC} & Lung Nodule & CT & \multicolumn{2}{l}{LUNA~2016~\citep{setio2017validation}} & Lung nodule false positive reduction \\
        \texttt{NCS} & Lung Nodule & CT & \multicolumn{2}{l}{LIDC-IDRI~\citep{armato2011lung}} & Lung nodule segmentation \\
        \texttt{ECC} & Pulmonary Emboli & CT & \multicolumn{2}{l}{PE-CAD~\citep{tajbakhsh2015computer}}  & Pulmonary embolism false positive reduction \\
        \texttt{LCS} & Liver & CT & \multicolumn{2}{l}{LiTS~2017~\citep{bilic2019liver}} & Liver segmentation \\
        \texttt{BMS} & Brain Tumor & MRI & \multicolumn{2}{l}{BraTS~2018~\citep{menze2015multimodal,bakas2018identifying}} & Brain tumor segmentation \\
        \shline
    \end{tabular}
    \begin{tablenotes}
        \item $^{\dagger}$ The first letter denotes the object of interest (``\texttt{N}'' for lung nodule, ``\texttt{E}'' for pulmonary embolism, ``\texttt{L}'' for liver, etc); the second letter denotes the modality (``\texttt{C}'' for CT, ``\texttt{M}'' for MRI, etc);  the last letter denotes the task (``\texttt{C}'' for classification, ``\texttt{S}'' for segmentation).
    \end{tablenotes}
\end{threeparttable}
\end{table*}

\begin{figure*}[t]
\begin{center}
\includegraphics[width=1.0\linewidth]{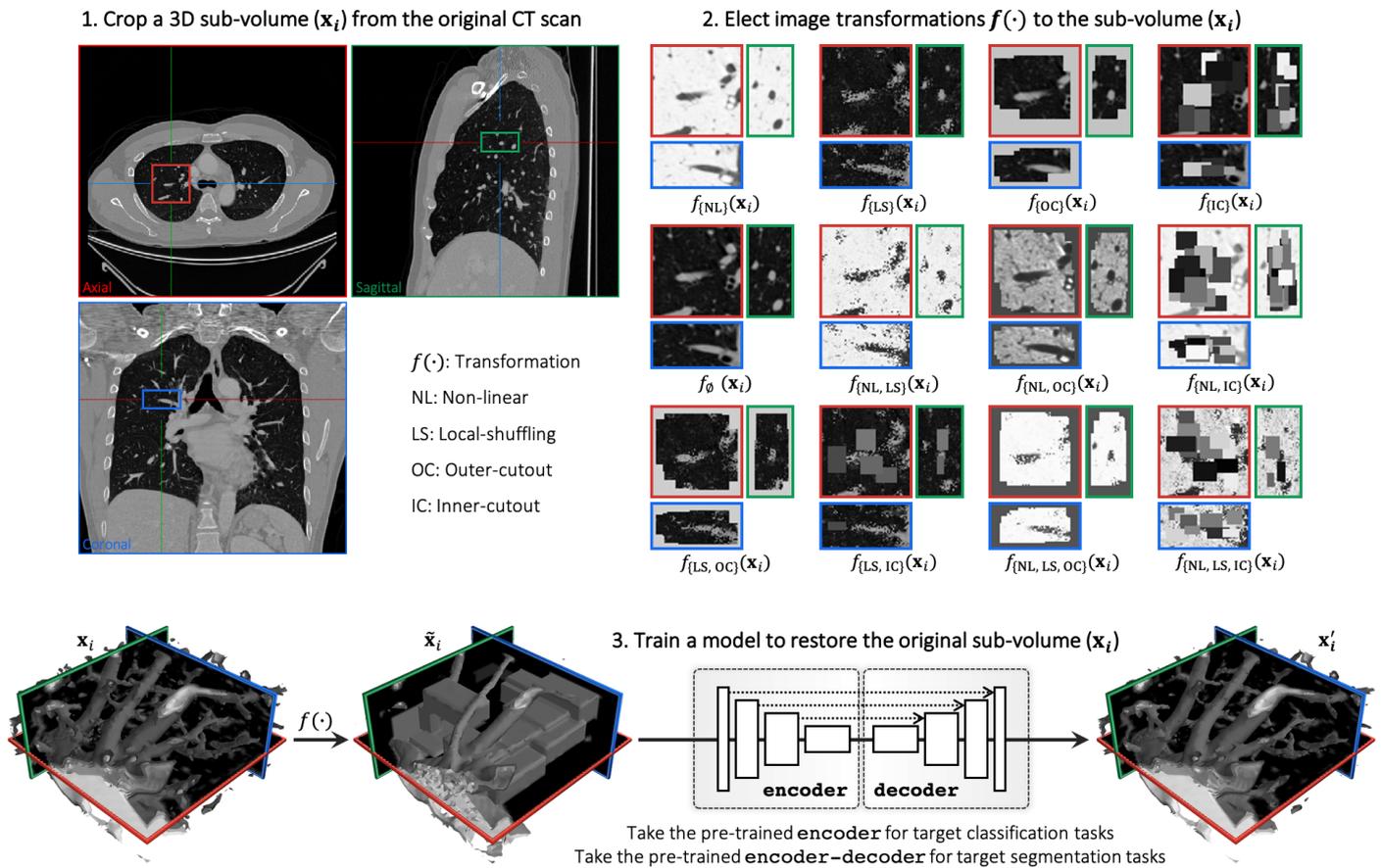}
\end{center}
\caption{
[Better viewed on-line, in color, and zoomed in for details] Our self-supervised learning framework aims to learn general-purpose image representation by recovering the original sub-volumes of images from their transformed ones. We first crop arbitrarily-size sub-volume $\textbf{x}_i$ at a random location from an unlabeled CT image. Each sub-volume $\textbf{x}_i$ can undergo at most three out of four transformations: non-linear, local-shuffling, outer-cutout, and inner-cutout, resulting in a transformed sub-volume $\tilde{\textbf{x}}_i$. It should be noted that outer-cutout and inner-cutout are considered mutually exclusive. Therefore, in addition to the four original individual transformations, this process yields eight more transformations, including one identity mapping ($\phi$ meaning none of the four individual transformations is selected) and seven combined transformations. A Model Genesis, an encoder-decoder architecture with skip connections in between, is trained to learn a common image representation by restoring the original sub-volume $\textbf{x}_i$ (as ground truth) from the transformed one $\tilde{\textbf{x}}_i$ (as input), in which the reconstruction loss (MSE) is computed between the model prediction $\textbf{x}_i'$ and ground truth $\textbf{x}_i$. Once trained, the encoder alone can be fine-tuned for target classification tasks; while the encoder and decoder together can be fine-tuned for target segmentation tasks. 
}
\label{fig:self_supervised_learning_framework}
\end{figure*}

Transfer learning from {\em natural} images to {\em medical} images has become the \textit{de facto} standard in deep learning for medical image analysis~\citep{tajbakhsh2016convolutional,shin2016deep}, but given the marked differences between {\em natural} images and {\em medical} images, we hypothesize that transfer learning can yield more powerful (application-specific) {\em target} models from the {\em source} models built directly using medical images. To test this hypothesis, we have chosen chest imaging because the chest contains several critical organs, which are prone to a number of diseases that result in substantial morbidity and mortality, hence associated with significant health-care costs. In this research, we focus on Chest CT, because of its prominent role in diagnosing lung diseases, and our research community has accumulated several Chest CT image databases, for instance, LIDC-IDRI~\citep{armato2011lung} and NLST~\citep{national2011reduced}, containing a large number of Chest CT images. However, systematically annotating Chest CT scans is not only tedious, laborious, and time-consuming, but it also demands costly, specialty-oriented skills, which are not easily accessible. Therefore, we seek to answer the following question: {\em Can we utilize the large number of available Chest CT images without systematic annotation to train source models that can yield high-performance target models via transfer learning?} 

To answer this question, we have developed a framework that trains generic source models for 3D medical imaging. Our framework is {\em autodidactic}---eliminating the need for labeled data by self-supervision; {\em robust}---learning comprehensive image representation from a mixture of self-supervised tasks; {\em scalable}---consolidating a variety of self-supervised tasks into a single image restoration task with the same encoder-decoder architecture; and {\em generic}---benefiting a range of 3D medical imaging tasks through transfer learning.
We call the models trained with our framework Generic Autodidactic Models, nicknamed Models Genesis, and refer to the model trained using Chest CT images as Genesis Chest CT.
As ablation studies, we have also trained a downgraded 2D version using 2D Chest CT slices, called Genesis Chest CT 2D. For thorough performance comparisons, we have trained a 2D model using Chest X-ray images, named as Genesis Chest X-ray (detailed in~\tablename~\ref{tab:terminology}). 

Naturally, 3D imaging tasks in the most prominent medical imaging modalities (\eg CT and MRI) should be solved directly in 3D, but 3D models generally have significantly more parameters than their 2D counterparts, thus demanding more labeled data for training. As a result, learning from scratch simply in 3D may {\em not} necessarily yield performance better than fine-tuning Models ImageNet (\ie pre-trained models on ImageNet), as revealed in \figurename~\ref{fig:2D_3D_target_tasks}. However, as demonstrated by our extensive experiments in Sec.~\ref{sec:experiments}, our Genesis Chest CT not only {\em significantly} outperforms learning 3D models from scratch (see \figurename~\ref{fig:random_initialization}), but also {\em consistently} tops any 2D/2.5D approaches including fine-tuning Models ImageNet as well as fine-tuning our Genesis Chest X-ray and Genesis Chest CT 2D (see \figurename~\ref{fig:2D_3D_target_tasks} and \tablename~\ref{tab:3d_2.5d_2d}). Furthermore, Genesis Chest CT surpasses publicly-available, pre-trained, (fully) supervised 3D models (see \tablename~\ref{tab:top_existing_models}). Our results  confirm the importance of 3D anatomical information and demonstrate the significance of Models Genesis for 3D medical imaging.

This performance is attributable to the following key observation: medical imaging protocols typically focus on particular parts of the body for specific clinical purposes, resulting in images of similar anatomy. The sophisticated yet recurrent anatomy offers consistent patterns for self-supervised learning to discover common representation of a particular body part (the lungs in our case). 
As illustrated in \figurename~\ref{fig:self_supervised_learning_framework}, the fundamental idea behind our self-supervised learning method is to recover anatomical patterns from images transformed via various ways in a unified framework. 

In summary, we make the following three contributions:
\begin{enumerate}
    \item A collection of generic pre-trained 3D models, performing effectively across diseases, organs, and modalities.
    \item A scalable self-supervised learning framework, offering encoder for classification and encoder-decoder for segmentation.
    \item A set of self-supervised training schemes, learning robust representation from multiple perspectives.
\end{enumerate}

In the remainder of this paper, we first in Sec.~\ref{sec:method} introduce our self-supervised learning framework for training Models Genesis, covering our four proposed image transformations with their learning perspectives, and describing the four unique properties of our Models Genesis. Sec.~\ref{sec:experiments} details the training process of Models Genesis and the five target tasks for evaluating Models Genesis, while Sec.~\ref{sec:results} summarizes the five major observations from our extensive experiments, demonstrating that our Models Genesis can serve as a primary source of transfer learning for 3D medical imaging. In Sec.~\ref{sec:discussion}, we discuss various aspects of Models Genesis, including their relationship with automated data augmentation, their impact on the creation of a medical ImageNet, and their capabilities for same- and cross-domain transfer learning followed by a thorough review of existing supervised and self-supervised representation learning approaches in medical imaging in Sec.~\ref{sec:related_works}. Finally, Sec.~\ref{sec:conclusion} concludes and outlines future extensions of Models Genesis.

\begin{figure*}[t]
\begin{center}
\includegraphics[width=1.0\linewidth]{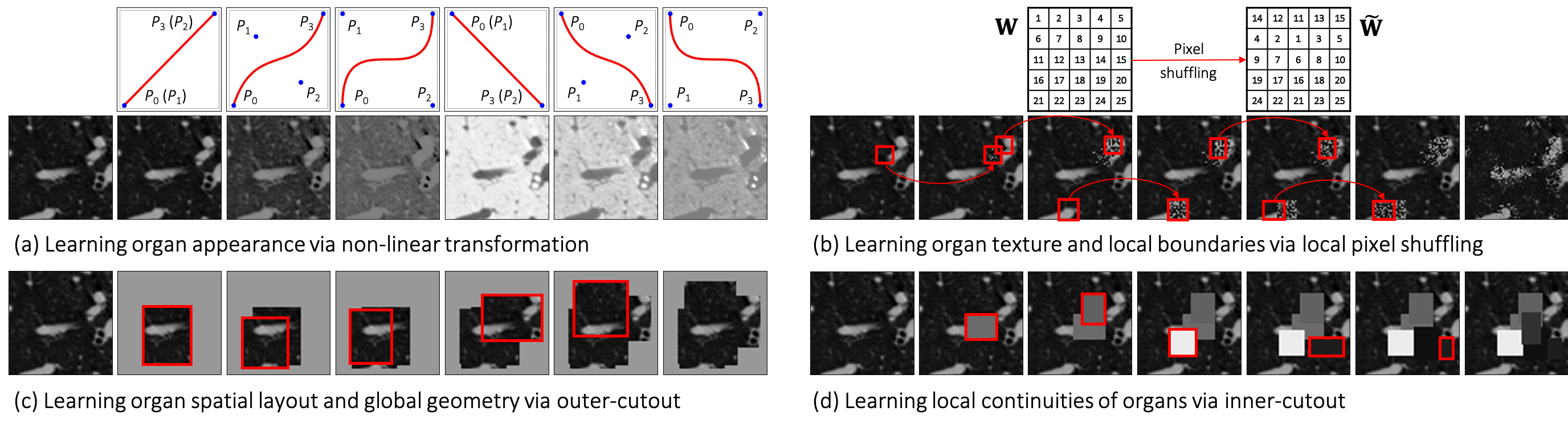}
\end{center}
\caption{
Illustration of the proposed image transformations and their learning perspectives. For simplicity and clarity, we illustrate the transformation on a 2D CT slice, but our Genesis Chest CT is trained directly using 3D sub-volumes, which are transformed in a 3D manner. For ease of understanding, in (a) non-linear transformation, we have displayed an image undergoing different translating functions in Columns 2---7; in (b) local-shuffling, (c) outer-cutout, and (d) inner-cutout transformation, we have illustrated each of the processes step by step in Columns 2---6, where the first and last columns denote the original images and the final transformed images, respectively. In local-shuffling, a different window $\mathbf{W}$ is automatically generated and used in each step. We provide the implementation details in Sec.~\ref{sec:image_transformation} and more visualizations in~\figurename~\ref{fig:transformation_appendix}.
}
\label{fig:image_transformations}
\end{figure*}

\section{Models Genesis}
\label{sec:method}

The objective of Models Genesis is to learn a common image representation that is transferable and generalizable across diseases, organs, and modalities. 
\figurename~\ref{fig:self_supervised_learning_framework} depicts our self-supervised learning framework, which enables training 3D models from scratch using unlabeled images, consisting of three steps: (1) cropping sub-volumes from patient CT images, (2) deforming the sub-volumes, and (3) training a model to restore the original sub-volume. In the following sections, we first introduce the denotations of our self-supervised learning framework and then detail each of the training schemes with its learning objectives and perspectives, followed by a summary of the four unique properties of our Models Genesis.

\subsection{Models Genesis learn by image restoration}

Given a raw dataset consisting of $N$ patient volumes, theoretically we can crop infinite number of sub-volumes from the dataset. In practice, we randomly generate a subset $\mathcal{X}=\{\mathbf{x_1},\mathbf{x_2},...,\mathbf{x_n}\}$, which includes $n$ number of sub-volumes and then apply image transformation function to these sub-volumes, yielding
\begin{equation}
    \tilde{\mathcal{X}} = f(\mathcal{X}),
\end{equation}
where $\tilde{\mathcal{X}}=\{\mathbf{\tilde{x}_1},\mathbf{\tilde{x}_2},...,\mathbf{\tilde{x}_n}\}$ and $f(\cdot)$ denotes a transformation function. Subsequently, a Model Genesis, being an encoder-decoder network with skip connections in between, will learn to approximate the function $g(\cdot)$ which aims to map the transformed sub-volumes $\tilde{\mathcal{X}}$ back to their original ones $\mathcal{X}$, that is,
\begin{equation}
    g(\tilde{\mathcal{X}}) = \mathcal{X} =  f^{-1}(\tilde{\mathcal{X}}).
\end{equation}

To avoid heavy weight dedicated towers for each proxy task and to maximize parameter sharing in Models Genesis, we consolidate four self-supervised schemes into a single image restoration task, enabling models to learn robust image representation by restoring from various sets of image transformations. Our proposed framework includes four transformations: (1) non-linear, (2) local-shuffling, (3) outer-cutout, and (4) inner-cutout.
Each transformation is independently applied to a sub-volume with a predefined probability, while outer-cutout and inner-cutout are considered mutually exclusive. Consequently, each sub-volume can undergo at most three of the above transformations, resulting in twelve possible transformed sub-volume (see step 2 in~\figurename~\ref{fig:self_supervised_learning_framework}). For clarity, we further define a {\em training scheme} as the process that (1) transforms sub-volumes using any of the aforementioned transformations, and (2) trains a model to restore the original sub-volumes from the transformed ones. For convenience, we refer to an {\em individual training scheme} as the scheme using one particular individual transformation.
We should emphasize that our ultimate goal is not the task of image restoration \textit{per se}. While restoring images is advocated and investigated as a training scheme for models to learn image representation, the usefulness of the learned representation must be assessed \textit{objectively} based on its generalizability and transferability to various target tasks.

\subsection{Models Genesis learn from multiple perspectives}
\label{sec:image_transformation}

\medskip
\noindent{\bf 1) Learning appearance via non-linear transformation.} 
We propose a novel self-supervised training scheme based on non-linear translation, with which the model learns to restore the intensity values of an input image transformed with a set of non-linear functions. The rationale is that the absolute intensity values (\ie Hounsfield units) in CT scans or relative intensity values in other imaging modalities convey important information about the underlying structures and organs~\citep{buzug2011computed,forbes2012human}. Hence, this training scheme enables the model to learn the appearance of the anatomic structures present in the images. In order to keep the appearance of the anatomic structures perceivable, we intentionally retain the non-linear intensity transformation function as \textit{monotonic}, allowing pixels of different values to be assigned with new distinct values. To realize this idea, we use B{\'e}zier Curve~\citep{mortenson1999mathematics}, a smooth and monotonic transformation function, which is generated from two end points ($P_0$ and $P_3$) and two control points ($P_1$ and $P_2$), defined as:
\begin{equation}
    B(t)=(1-t)^3P_0+3(1-t)^2tP_1+3(1-t)t^2P_2+t^3P_3,\ t\in [0,1],
\end{equation}
where $t$ is a fractional value along the length of the line. In \figurename~\ref{fig:image_transformations}(a), we illustrate the original CT sub-volume (the left-most column) and its transformed ones based on different transformation functions. The corresponding transformation functions are shown in the top row. Notice that, when $P_0 = P_1$ and $P_2 = P_3$ the B{\'e}zier Curve is a linear function (shown in Columns 2, 5). Besides, we set $P_0 = (0,0)$ and $P_3 = (1,1)$ to get an increasing function (shown in Columns 2---4) and the opposite to get a decreasing function (shown in Columns 5---7). The control points are randomly generated for more variances (shown in Columns 3, 4, 6, 7). Before applying the transformation functions, in Genesis CT, we first clip the Hounsfield units values within the range of $[-1000, 1000]$ and then normalize each CT scan to $[0, 1]$, while in Genesis X-ray, we directly normalize each X-ray to $[0, 1]$ without intensity clipping.
     
\medskip
\noindent{\bf 2) Learning texture via local pixel shuffling.} We propose local pixel shuffling to enrich local variations of a sub-volume without dramatically compromising its global structures, which encourages the model to learn the {\em local} boundaries and textures of objects. To be specific, for each input sub-volume, we randomly select 1,000 windows and then shuffle the pixels inside each window sequentially. Mathematically, let us consider a small window $\mathbf{W}$ with a size of $m\times n$.
The local-shuffling acts on each window and can be formulated as
\begin{equation}
    \tilde{\mathbf{W}}=\mathbf{P}\times\mathbf{W}\times\mathbf{P}',
\end{equation}
where $\tilde{\mathbf{W}}$ is the transformed window, $\mathbf{P}$ and $\mathbf{P}'$ denote permutation metrics with the size of $m\times m$ and $n\times n$, respectively. Pre-multiplying $\mathbf{W}$ with $\mathbf{P}$ permutes the rows of the window $\mathbf{W}$, whereas post-multiplying $\mathbf{W}$ with $\mathbf{P}'$ results in the permutation of the columns of the window $\mathbf{W}$. The size of the local window determines the difficulty of proxy task. In practice, to preserve the global content of the image, we keep the window sizes smaller than the receptive field of the network, so that the network can learn much more robust image representation by ``resetting'' the original pixels positions. Note that our method is quite different from PatchShuffling~\citep{kang2017patchshuffle}, which is a regularization technique to avoid over-fitting. Unlike de-noising~\citep{vincent2010stacked} and in-painting~\citep{pathak2016context,iizuka2017globally}, our local-shuffling transformation does not intend to replace the pixel values with noise, which therefore preserves the identical global distributions to the original sub-volume. 
In addition, local-shuffling within an extent keeps the objects perceivable, as shown in \figurename~\ref{fig:image_transformations}(b), benefiting the deep neural network in learning \textit{local} invariant image representations, which serves as a complementary perspective with global patch shuffling~\citep{chen2019self} (discussed in-depth in \ref{sec:localshuffling_patchshuffling}).

\medskip
\noindent{\bf 3) Learning context via outer and inner cutouts.} We devise outer-cutout as a new training scheme for self-supervised learning. To realize it, we generate an arbitrary number ($\leq 10$) of windows, with various sizes and aspect ratios, and superimpose them on top of each other, resulting in a single window of a complex shape. When applying this merged window to a sub-volume, we leave the sub-volume region inside the window exposed and mask its surrounding (\ie outer-cutout) with a random number. 
Moreover, to prevent the task from being too difficult or even unsolvable, we extensively search for the optimal size of cutout regions spanning from 0\% to 90\%, incremented by 10\% (detailed study presented in \ref{sec:cutout_region_study}). In the end, we limit the outer-cutout region to be less than 1/4 of the whole sub-volume.
By restoring the outer-cutouts, the model will learn the {\em global} geometry and spatial layout of organs in medical images via extrapolating within each sub-volume. We have illustrated this process step by step in \figurename~\ref{fig:image_transformations}(c). The first and last columns denote the original sub-volumes and the final transformed sub-volumes, respectively. 

Our self-supervised learning framework also utilizes inner-cutout as a training scheme, where we mask the inner window regions (\ie inner-cutouts) and leave their surroundings exposed. By restoring the inner-cutouts, the model will learn {\em local} continuities of organs in medical images via interpolating within each sub-volume. Unlike \citet{pathak2016context}, where in-painting is proposed as a proxy task by restoring only the central region of the image, we restore the entire sub-volume as the model output. Examples of inner-cutout are illustrated in \figurename~\ref{fig:image_transformations}(d). 
Following the suggestion from~\citet{pathak2016context}, the inner-cutout areas are limited to be less than $1/4$ of the whole sub-volume, in order to keep the task reasonably difficult.

\subsection{Models Genesis have several unique properties}
\label{sec:properties}

\medskip
\noindent{\bf 1) Autodidactic---requiring no manual labeling.} Models Genesis are trained in a self-supervised manner with abundant unlabeled image datasets, demanding {\em zero} expert annotation effort. Consequently, Models Genesis are fundamentally different from traditional (fully) {\em supervised} transfer learning from ImageNet~\citep{shin2016deep,tajbakhsh2016convolutional}, which offers modest benefits to 3D medical imaging applications as well as that from the existing pre-trained, full-supervised models including I3D~\citep{carreira2017quo}, NiftyNet~\citep{gibson2018niftynet}, and MedicalNet~\citep{chen2019med3d}, which demand a volume of annotation effort to obtain the source models (statistics given in \tablename~\ref{tab:terminology}).
To our best knowledge, this work represents the first effort to establish publicly-available, autodidactic models for 3D medical image analysis.

\medskip
\noindent{\bf 2) Robust---learning from multiple perspectives.} Our combined approach trains Models Genesis from multiple perspectives (appearance, texture, context, etc.), leading to more robust models across all target tasks, as evidenced in \figureautorefname~\ref{fig:combined_vs_individuals}, where our combined approach is compared with our individual schemes. This eclectic approach, incorporating multiple tasks into a single image restoration task, empowers Models Genesis to learn more comprehensive representation. 
While most self-supervised methods devise isolated training schemes to learn from specific perspectives---learning intensity value via colorization, context information via Jigsaw, orientation via rotation, etc---these methods are reported with mixed results on different tasks, in review papers such as~\citet{goyal2019scaling},~\citet{kolesnikov2019revisiting},~\citet{taleb20203d}, and~\citet{jing2020self}. It is critical as a multitude of state-of-the-art results in the literature show the importance of using compositions of more than one transformations per image~\citep{graham2014fractional,dosovitskiy2015discriminative,wu2020generalization}, which has also been experimentally confirmed in our image restoration task.

\medskip
\noindent{\bf 3) Scalable---accommodating many training schemes.}
Consolidated into a single image restoration task, our novel self-supervised schemes share the same encoder and decoder during training. Had each task required its own decoder, due to limited memory on GPUs, our framework would have failed to accommodate a large number of self-supervised tasks. By unifying all tasks as a single image restoration task, any favorable transformation can be easily amended into our framework, overcoming the scalability issue associated with multi-task learning~\citep{doersch2017multi,noroozi2018boosting,standley2019tasks,chen2019med3d}, where the network heads are subject to the specific proxy tasks.

\medskip
\noindent{\bf 4) Generic---yielding diverse applications.}
Models Genesis, trained via a diverse set of self-supervised schemes, learn a general-purpose image representation that can be leveraged for a wide range of target tasks. Specifically, Models Genesis can be utilized to initialize the encoder for the target {\em classification} tasks and to initialize the encoder-decoder for the target {\em segmentation} tasks, while the existing self-supervised approaches are largely focused on providing encoder models only~\citep{jing2020self}. 
As shown in \tablename~\ref{tab:top_existing_models}, Models Genesis can be generalized across diseases (\eg nodule, embolism, tumor), organs (\eg lung, liver, brain), and modalities (\eg CT and MRI), a generic behavior that sets us apart from all previous works in the literature where the  representation is learned via a specific self-supervised task, and thus lack generality.

\section{Experiments}
\label{sec:experiments}

\subsection{Pre-training Models Genesis}
\label{sec:experiments_pretraining_models_genesis}

Our Models Genesis are pre-trained from 623 Chest CT scans in LUNA~2016~\citep{setio2017validation} in a self-supervised manner. The reason that we decided not to use all 888 scans provided by this dataset was to avoid test-image leaks between proxy and target tasks, so that we can confidently use the rest of the images solely for testing Models Genesis as well as the target models, although Models Genesis are trained from only unlabeled images, involving no annotation shipped with the dataset. We first randomly crop sub-volumes, sized $64\times 64\times 32$ pixels, from different locations. To extract more informative sub-volumes for training, we then intentionally exclude those which are empty (air) or contain full tissues. Our Models Genesis 2D are self-supervised pre-trained from LUNA~2016~\citep{setio2017validation} and ChestX-ray14~\citep{wang2017chestx} using 2D CT slices in an axial view and X-ray images, respectively. For all proxy tasks and target tasks, the raw image intensities were normalized to the $[0,1]$ range before training. We use the mean square error (MSE) between input and output images as objective function for the proxy task of image restoration. As suggested by~\citet{pathak2016context} and \citet{chen2019self}, the MSE loss is sufficient for representation learning, although the restored images may be blurry.

When pre-training Models Genesis, we apply each of the transformations on sub-volumes with a pre-defined probability. That being said, the model will encounter not only the transformed sub-volumes as input, but also the original sub-volumes. This design offers two advantages:
\begin{itemize}
    \item First, the model must distinguish original versus transformed images, discriminate transformation type(s), and restore images if transformed. Our self-supervised learning framework, therefore, results in pre-trained models that are capable of handling versatile tasks.
    \item Second, since original images are presented in the proxy task, the semantic difference of input images between the proxy and target task becomes smaller. As a result, the pre-trained model can be transferable to process regular/normal images in a broad variety of target tasks.
\end{itemize}

\begin{table}[t]
\centering
\footnotesize
\caption{
    Genesis CT is pre-trained on \textit{only} LUNA~2016 dataset (\ie the source) and then fine-tuned for five distinct medical image applications (\ie the targets). These target tasks are selected such that they show varying levels of semantic distance from the source, in terms of organs, diseases, and modalities, allowing us to investigate the transferability of the pre-trained weights of Genesis CT with respect to the domain distance. The cells checked by \xmark \ denote the properties that are different between the source and target datasets.
}
\label{tab:distance}
\begin{tabular}{p{0.15\linewidth}P{0.15\linewidth}P{0.15\linewidth}P{0.15\linewidth}P{0.15\linewidth}}
    \shline
    Task & Disease & Organ & Dataset & Modality \\
    \hline
    \texttt{NCC} &  \\
    \texttt{NCS} &  \\
    \texttt{ECC} & \xmark & & \xmark \\
    \texttt{LCS} & \xmark & \xmark & \xmark \\
    \texttt{BMS} & \xmark & \xmark & \xmark & \xmark \\
    \shline
\end{tabular}
\end{table}

\subsection{Fine-tuning Models Genesis}
\label{sec:finetuning_models_genesis}

The pre-trained Models Genesis can be adapted to new imaging tasks through transfer learning or fine-tuning. There are three major transfer learning scenarios: (1) employing the encoder as a fixed feature extractor for a new dataset and following up with a linear classifier (\eg Linear SVM or Softmax classifier), (2) taking the pre-trained encoder and appending a sequence of fully-connected (\textit{fc}) layers for target classification tasks, and (3) taking the pre-trained encoder and decoder and replacing the last layer with a $1\times 1\times 1$ convolutional layer for target segmentation tasks. For scenarios (2) and (3), it is possible to fine-tune all the layers of the model or to keep some of the earlier layers fixed, only fine-tuning some higher-level portion of the model. We have evaluated the performance of our self-supervised representation for transfer learning by fine-tuning all layers in the network. In the following, we examine Models Genesis on five distinct medical applications, covering classification and segmentation tasks in CT and MRI images with varying levels of semantic distance from the source (Chest CT) to the targets in terms of organs, diseases, and modalities (see \tablename~\ref{tab:distance}) for investigating the transferability of Models Genesis.

\subsubsection{Lung nodule false positive reduction (\texttt{NCC})}
The dataset is provided by LUNA~2016~\citep{setio2017validation} and consists of 888 low-dose lung CTs with slice thickness less than 2.5mm. Patients are randomly assigned into a training set (445 cases), a validation set (178 cases), and a test set (265 cases). The dataset offers the annotations for a set of 5,510,166 candidate locations for the false positive reduction task, wherein true positives are labeled as ``1'' and false positives are labeled as ``0''. Following the prior works~\citep{setio2016pulmonary,sun2017automatic}, we evaluate performance via Area Under the Curve (AUC) score on classifying true positives and false positives.

\subsubsection{Lung nodule segmentation (\texttt{NCS})}
The dataset is provided by the Lung Image Database Consortium image collection (LIDC-IDRI)~\citep{armato2011lung} and consists of 1,018 cases collected by seven academic centers and eight medical imaging companies. The cases were split into training (510), validation (100), and test (408) sets. Each case is a 3D CT scan and the nodules have been marked as volumetric binary masks. We have re-sampled the volumes to 1-1-1 spacing and then extracted a $64\times 64\times 32$ crop around each nodule. These 3D crops are used for model training and evaluation. As in prior works~\citep{aresta2019iw,tang2019nodulenet,zhou2018unet++}, we adopt Intersection over Union (IoU) and Dice coefficient scores to evaluate performance. Note that for this particular application, we calculate mean of the IoUs at thresholds ranging from 0.5 to 0.95 with a step size of 0.05.

\subsubsection{Pulmonary embolism false positive reduction (\texttt{ECC})}
We utilize a database consisting of 121 computed tomography pulmonary angiography (CTPA) scans with a total of 326 emboli. 
Following the prior works~\citep{liang2007computer}, we utilize their PE candidate generator based on the toboggan algorithm, resulting in total of 687 true positives and 5,568 false positives. The dataset is then divided at the patient-level into a training set with 434 true positive PE candidates and 3,406 false positive PE candidates, and a test set with 253 true positive PE candidates and 2,162 false positive PE candidates. To conduct a fair comparison with the prior study~\citep{zhou2017fine,tajbakhsh2016convolutional,tajbakhsh2019computer}, we compute candidate-level AUC on classifying true positives and false positives.

\subsubsection{Liver segmentation (\texttt{LCS})}
The dataset is provided by MICCAI 2017 LiTS Challenge and consists of 130 labeled CT scans, which we split into training (100 patients), validation (15 patients), and test (15 patients) subsets. The ground truth segmentation provides two different labels: liver and lesion. For our experiments, we only consider liver as positive class and others as negative class and evaluate segmentation performance using Intersection over Union (IoU) and Dice coefficient scores.

\subsubsection{Brain tumor segmentation (\texttt{BMS})} 
The dataset is provided by BraTS~2018 challenge~\citep{menze2015multimodal,bakas2018identifying} and consists of 285 patients (210 HGG and 75 LGG), each with four 3D MRI modalities (T1, T1c, T2, and Flair) rigidly aligned. We adopt 3-fold cross validation, in which two folds (190 patients) are for training and one fold (95 patients) for test. Annotations include background (label 0) and three tumor subregions: GD-enhancing tumor (label 4), the peritumoral edema (label 2), and the necrotic and non-enhancing tumor core (label 1). We consider those with label 0 as negatives and others as positives and evaluate segmentation performance using Intersection over Union (IoU) and Dice coefficient scores.

\subsection{Benchmarking Models Genesis}
\label{sec:baselines_implementation}

For a thorough comparison, we used three different techniques to randomly initialize the weights of models: (1) a basic random initialization method based on Gaussian distributions, (2) a method commonly known as Xavier, which was suggested in~\citet{glorot2010understanding}, and (3) a revised version of Xavier called MSRA, which was suggested in~\citet{he2015delving}. They are implemented as \texttt{uniform}, \texttt{glorot\_uniform}, and \texttt{he\_uniform}, respectively, following the Initializers\footnote{\label{foot:Initializer}Initializers: \href{http://faroit.com/keras-docs/1.2.2/initializations/}{faroit.com/keras-docs/1.2.2/initializations}} in Keras. We compare Models Genesis with Rubik's cube~\citep{zhuang2019self}, the most recent multi-task and self-supervised learning method for 3D medical imaging. Considering that most of the self-supervised learning methods are initially proposed and implemented in 2D, we have extended five most representative ones~\citep{vincent2010stacked,pathak2016context,noroozi2016unsupervised,chen2019self,caron2018deep} into their 3D versions for a fair comparison (see detailed implementation in~\ref{sec:baseline_implementation_appendix}). To promote the 3D self-supervised learning research, we make our own implementation of the 3D extended methods and their corresponding pre-trained weights publicly available as an open-source tool that can effectively be used out-of-the-box. In addition, we have examined publicly available pre-trained models for 3D transfer learning in medical imaging, including NiftyNet\footnote{\label{foot:NiftyNet}NiftyNet Model Zoo: \href{https://github.com/NifTK/NiftyNetModelZoo}{github.com/NifTK/NiftyNetModelZoo}}~\citep{gibson2018niftynet}, MedicalNet\footnote{\label{foot:MedicalNet}MedicalNet: \href{https://github.com/Tencent/MedicalNet}{github.com/Tencent/MedicalNet}}~\citep{chen2019med3d}, and, the most influential 2D weights initialization, Models ImageNet. We also fine-tune I3D\footnote{\label{foot:I3D}I3D: \href{https://github.com/deepmind/kinetics-i3d}{github.com/deepmind/kinetics-i3d}}~\citep{carreira2017quo} in our five target tasks because it has been shown to successfully initialize 3D models for lung nodule detection in~\citet{ardila2019end}. The detailed configurations of these models can be found in \ref{sec:public_3d_model_appendix}.

3D U-Net architecture\footnote{\label{foot:3dunet}3D U-Net: \href{https://github.com/ellisdg/3DUnetCNN}{github.com/ellisdg/3DUnetCNN}} is used in 3D applications; U-Net architecture\footnote{\label{foot:densenet121}Segmentation Models: \href{https://github.com/qubvel/segmentation_models}{github.com/qubvel/segmentation\_models}} is used in 2D applications. Batch normalization~\citep{ioffe2015batch} is utilized in all 3D/2D deep models. For proxy tasks, SGD method~\citep{zhang2004solving} with an initial learning rate of $1e0$ is used for optimization. We use \texttt{ReduceLROnPlateau} to schedule learning rate, in which if no improvement is seen in the validation set for a certain number of epochs, the learning rate is reduced. For target tasks, Adam method~\citep{kinga2015method} with a learning rate of $1e-3$ is used for optimization, where $\beta_1=0.9$, $\beta_2=0.999$, $\epsilon=1e-8$. We use \textit{early-stop} mechanism on the validation set to avoid over-fitting. Simple yet heavy 3D data augmentation techniques are employed in all five target tasks, including random flipping, transposing, rotating, and adding Gaussian noise. We run each method ten times on all of the target tasks and report the average, standard deviation, and further present statistical analysis based on an independent two-sample \textit{t}-test.

In the proxy task, we pre-train the model using 3D sub-volumes sized $64\times 64\times 32$, whereas in target tasks, the input is not limited to sub-volumes with certain size. That being said, our pre-trained models can be fine-tuned in the tasks with CT sub-volumes, entire CT volumes, or even MRI volumes as input upon user's need. The flexibility of input size is attributed to two reasons: (1) our pre-trained models learn generic image representation such as appearance, texture, and context feature, and (2) the encoder-decoder architecture is able to process images with arbitrary sizes.

\begin{figure*}[t]
\begin{center}
\includegraphics[width=1.0\linewidth]{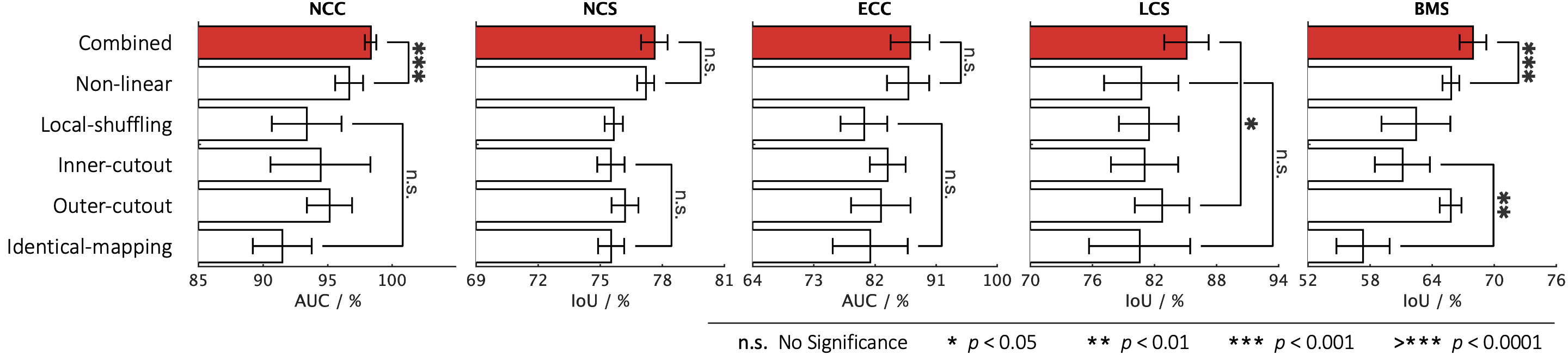}
\end{center}
\caption{
Comparing the combined training scheme with each of the proposed individual training schemes, we conduct statistical analyses between the top two training schemes as well as between the bottom two. Although some of the individual training schemes could be favorable for certain target tasks, there is no such clear clue to guarantee that any one of the individual training schemes would consistently offer the best performance on every target task. On the contrary, our combined training scheme consistently achieves the best results across all five target tasks.
}
\label{fig:combined_vs_individuals}
\end{figure*}

\begin{figure*}[t]
\begin{center}
\includegraphics[width=1.0\linewidth]{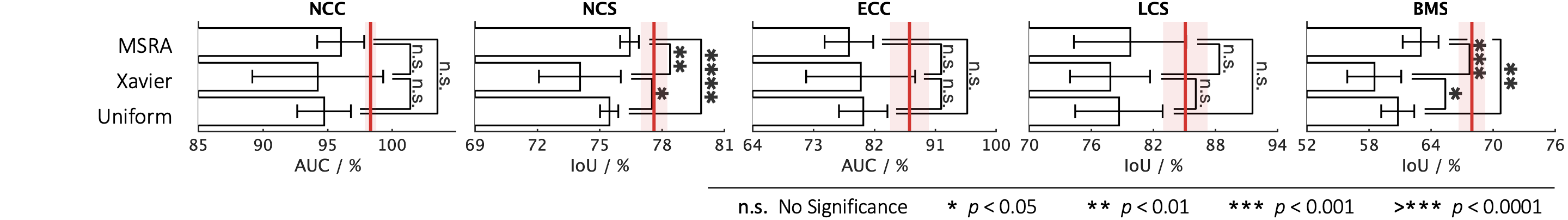}
\end{center}
\caption{
Models Genesis, as presented with the red vertical lines, achieve higher and more stable performance compared with three popular types of random initialization methods, including MSRA, Xavier, and Uniform. Among three out of the five applications, three different types of random distribution reveal no significant difference with respect to each other.
}
\label{fig:random_initialization}
\end{figure*}

\begin{figure*}[t]
\begin{center}
\includegraphics[width=1.0\linewidth]{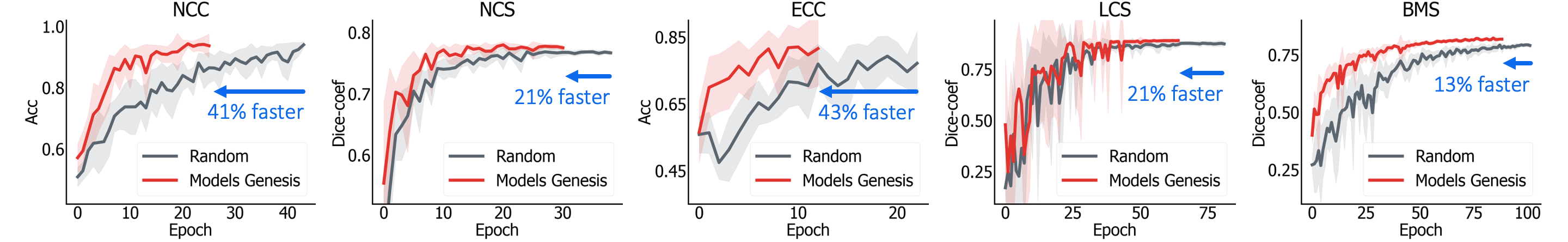}
\end{center}
\caption{
Models Genesis enable better optimization than learning from scratch, evident by the learning curves for the target tasks of reducing false positives in detecting lung nodules (\texttt{NCC}) and pulmonary embolism (\texttt{ECC}) as well as segmenting lung nodule (\texttt{NCS}), liver (\texttt{LCS}), and brain tumor (\texttt{BMS}). We have plotted the validation performance averaged by ten trials for each application, in which accuracy and dice-coefficient scores are reported for classification and segmentation tasks, respectively. As seen, initializing with our pre-trained Models Genesis demonstrates benefits in the convergence speed.
}
\label{fig:learning_curve}
\end{figure*}

\begin{table*}[t]
\centering
\footnotesize
\caption{
Our Models Genesis lead the best or comparable performance on five distinct medical target tasks over six self-supervised learning approaches (revised in 3D) and three competing publicly available (fully) supervised pre-trained 3D models. For ease of comparison, we evaluate AUC score for the two classification tasks (\ie \texttt{NCC} and \texttt{ECC}) and IoU score for the three segmentation tasks (\ie \texttt{NCS}, \texttt{LCS}, and \texttt{BMS}). All of the results, including the mean and standard deviation (mean$\pm$s.d.) across ten trials, reported in the table are evaluated using our dataset splitting, elaborated in~Sec.~\ref{sec:finetuning_models_genesis}. For every target task, we have further performed independent two sample $t$-test between the best (bolded) vs. others and highlighted boxes in blue when they are not statistically significantly different at $p=0.05$ level. The footnotes compare our results with the state-of-the-art performance for each target task, using the evaluation metric for the data acquired from competitions. 
}
\label{tab:top_existing_models}
\begin{tabular}{p{0.12\linewidth}p{0.32\linewidth}P{0.08\linewidth}P{0.08\linewidth}P{0.08\linewidth}P{0.08\linewidth}P{0.08\linewidth}}
    \shline
    \multirow{2}{*}{Pre-training} & \multirow{2}{*}{Approach} & \multicolumn{5}{c}{Target tasks} \\
    \cline{3-7}
     & & \texttt{NCC}$^{1}$ ($\%$) & \texttt{NCS}$^{2}$ ($\%$) & \texttt{ECC}$^{3}$ ($\%$) & \texttt{LCS}$^{4}$ ($\%$) & \texttt{BMS}$^{5}$ ($\%$) \\
    \hline
    \multirow{3}{*}{No} & Random with Uniform Init & 94.74$\pm$1.97 & 75.48$\pm$0.43 & 80.36$\pm$3.58 & 78.68$\pm$4.23 & 60.79$\pm$1.60 \\
     & Random with Xavier Init~\citep{glorot2010understanding} & 94.25$\pm$5.07 & 74.05$\pm$1.97 & 79.99$\pm$8.06 & 77.82$\pm$3.87 & 58.52$\pm$2.61 \\
     & Random with MSRA Init~\citep{he2015delving} & 96.03$\pm$1.82 & 76.44$\pm$0.45 & 78.24$\pm$3.60 & 79.76$\pm$5.43 & 63.00$\pm$1.73 \\
    \hline
    \multirow{3}{*}{(Fully) supervised} & I3D~\citep{carreira2017quo} & \cellcolor{iblue!30}98.26$\pm$0.27 & 71.58$\pm$0.55 & 80.55$\pm$1.11 & 70.65$\pm$4.26 & \cellcolor{iblue!30}67.83$\pm$0.75 \\
     & NiftyNet~\citep{gibson2018niftynet} & 94.14$\pm$4.57 & 52.98$\pm$2.05 & 77.33$\pm$8.05 & 83.23$\pm$1.05 & 60.78$\pm$1.60  \\
     & MedicalNet~\citep{chen2019med3d} & 95.80$\pm$0.49 & 75.68$\pm$0.32 & \cellcolor{iblue!30}86.43$\pm$1.44 & \cellcolor{iblue!30}\textbf{85.52$\pm$0.58}$^{\dagger}$ & 66.09$\pm$1.35 \\
    \hline
    \multirow{7}{*}{Self-supervised} & De-noising~\citep{vincent2010stacked} & 95.92$\pm$1.83 & 73.99$\pm$0.62 & \cellcolor{iblue!30}85.14$\pm$3.02 & 84.36$\pm$0.96 & 57.83$\pm$1.57 \\
     & In-painting~\citep{pathak2016context} & 91.46$\pm$2.97 & 76.02$\pm$0.55 & 79.79$\pm$3.55 & 81.36$\pm$4.83 & 61.38$\pm$3.84 \\
     & Jigsaw~\citep{noroozi2016unsupervised} & 95.47$\pm$1.24 & 70.90$\pm$1.55 & 81.79$\pm$1.04 & 82.04$\pm$1.26 & 63.33$\pm$1.11 \\
     & DeepCluster~\citep{caron2018deep} & 97.22$\pm$0.55 & 74.95$\pm$0.46 & 84.82$\pm$0.62 & 82.66$\pm$1.00 & 65.96$\pm$0.85 \\
     & Patch shuffling~\citep{chen2019self} & 91.93$\pm$2.32 & 75.74$\pm$0.51 & 82.15$\pm$3.30 & 82.82$\pm$2.35 & 52.95$\pm$6.92 \\
     & Rubik’s Cube~\citep{zhuang2019self} & 96.24$\pm$1.27 & 72.87$\pm$0.16 & 80.49$\pm$4.64 & 75.59$\pm$0.20 & 62.75$\pm$1.93 \\
     & Genesis Chest CT (ours) & \cellcolor{iblue!30}\textbf{98.34$\pm$0.44} & \cellcolor{iblue!30}\textbf{77.62$\pm$0.64} & \cellcolor{iblue!30}\textbf{87.20$\pm$2.87} & \cellcolor{iblue!30}85.10$\pm$2.15 & \cellcolor{iblue!30}\textbf{67.96$\pm$1.29} \\
    \shline
    \end{tabular}
    \begin{tablenotes}
        \item $^{1}$ The winner in~\citet{luna2016result} holds an official score of 0.968 vs. 0.971 (ours)
        \item $^{2}$ \citet{wu2018joint} holds a Dice of 74.05$\%$  vs. 75.86$\%\pm$0.90$\%$ (ours)
        \item $^{3}$ \citet{zhou2017fine} holds an AUC of 87.06$\%$ vs. 87.20$\%\pm$2.87$\%$ (ours)
        \item $^{4}$ The winner in~\citet{lits2017result} with post-processing holds a Dice of 96.60$\%$ vs. 93.19$\%\pm$0.46$\%$ (ours without post-processing)
        \item $^{5}$ MRI Flair images are only utilized for segmenting brain tumors, so the results are not submitted to BraTS~2018.
        \item $^{\dagger}$ Genesis Chest CT is slightly outperformed by MedicalNet in \texttt{LCS} because the latter has been (fully) supervised pre-trained on the LiTS dataset.
    \end{tablenotes}
\end{table*}

\section{Results}
\label{sec:results}

In this section, we begin with an ablation study to compare the combined approach with each individual scheme, concluding that the combined approach tends to achieve more robust results and consistently exceeds any other training schemes. We then take our pre-trained model from the combined approach and present results on five 3D medical applications, comparing them against the state-of-the-art approaches found in recent supervised and self-supervised learning literature.

\subsection{The combined learning scheme exceeds each individual}
\label{sec:individual_combination}

We have devised four individual training schemes by applying each of the transformations (\ie non-linear, local-shuffling, outer-cutout, and inner-cutout) individually to a sub-volume and training the model to restore the original one. We compare each of these training schemes with identical-mapping, which does not involve any image transformation. In three out of the five target tasks, as shown in Figs.~\ref{fig:combined_vs_individuals}---\ref{fig:random_initialization}, the model pre-trained by identical-mapping scheme does not perform as well as random initialization. This undesired representation obtained via identical-mapping suggests that without any image transformation, the model would not benefit much from the proxy image restoration task. On the contrary, nearly all of the individual schemes offer higher target task performances than identical-mapping, demonstrating the significance of the four devised image transformations in learning image representation. 

Although each of the individual schemes has established the capability in learning image representation, its empirical performance varies from task to task. That being said, given a target task, there is no clear winner among the four individual schemes that can always guarantee the highest performance. As a result, we have further devised a combined scheme, which applies transformations to a sub-volume with a predefined probability for each transformation and trains a model to restore the original one. To demonstrate the importance of combining these image transformations together, we examine the combined training scheme against each of the individual ones. \figurename~\ref{fig:combined_vs_individuals} shows that the combined scheme consistently exceeds any other individual schemes in all five target tasks. We have found that the combination of different transformations is advantageous because, as discussed, we cannot rely on one single training scheme to achieve the most robust and compelling results across multiple target tasks. 
It is our novel representation learning framework based on image restoration that allows integrating various training schemes into a single training scheme. Our qualitative assessment of image restoration quality, provided in~\figurename~\ref{fig:ct_restoration_individual_combined}, further indicates that the combined scheme is superior over all four individual schemes in restoring the images that have been undergone multiple transformations. In summary, our combined scheme pre-trains a model from multiple perspectives (appearance, texture, context, etc.), empowering models to learn a more comprehensive representation, thereby leading to more robust target models. Based on the above ablation studies, in the following sections, we refer the models pre-trained by the combined scheme to Models Genesis and, in particular, refer the model pre-trained on LUNA~2016 dataset to Genesis Chest CT.

\subsection{Models Genesis outperform learning from scratch}
\label{sec:surpass_scratch}

Transfer learning accelerates training and boosts performance, only if the image representation learned from the original (proxy) task is general and transferable to target tasks. Fine-tuning models trained on ImageNet has been a great success story in 2D~\citep{tajbakhsh2016convolutional,shin2016deep}, but for 3D representation learning, there is no such a massive labeled dataset like ImageNet. As a result, it is still common practice to train 3D model from scratch in 3D medical imaging.
Therefore, to establish the 3D baselines, we have trained 3D models with three representative random initialization methods, including naive uniform initialization, Xavier/Glorot initialization proposed by~\citet{glorot2010understanding}, and He normal (MSRA) initialization proposed by~\citet{he2015delving}. 
When comparing deep model initialization by transfer learning and by controlling mathematical distribution, the former learns more sophisticated image representation but suffers from a domain gap, whereas the latter is task independent yet provides relatively less benefit than the former. The hypothesis underneath transfer learning is that transferring deep features across visual tasks can obtain a semantically more powerful representation, compared with simply initializing weights using different distributions.
From our comprehensive experiments in~\figurename~\ref{fig:random_initialization}, we have observed the following:
\begin{itemize}
    \item Within each method, random initialization of weights has shown large variance in results of ten trials; it is in large part due to the difficulty of adequately initializing these networks from scratch. A small miscalibration of the initial weights can lead to vanishing or exploding gradients, as well as poor convergence properties. 
    \item In three out of the five 3D medical applications, the results reveal no significant difference among these random initialization methods. Although randomly initializing weights can vary by the behaviors on different applications, He normal (MSRA), in which the weights are initialized with a specific ReLU-aware initialization, generally works the most reliably among all five target tasks.
    \item On the other hand, initialization with our pre-trained Genesis Chest CT stabilizes the overall performance and, more importantly, elevates the average performance over all three random initialization methods by a large margin. Our statistical analysis shows that the performance gain is significant for all the target tasks under study. This suggests that, owing to the representation learning scheme, our initial weights provide a better starting point than the ones generated under particular statistical distributions, while being over 13\% faster (see \figurename~\ref{fig:learning_curve}). This observation has also been widely obtained in 2D model initialization~\citep{tajbakhsh2016convolutional,shin2016deep,rawat2017deep,zhou2017fine,voulodimos2018deep}.
\end{itemize}

Altogether, in contrast to 3D scratch models, we believe Models Genesis can potentially serve as a primary source of transfer learning for 3D medical imaging applications. Besides contrasting with the three random initialization methods, we further examine our Models Genesis against the existing pre-trained 3D models in the coming section.

\begin{figure*}[t]
\centering
\includegraphics[width=1.0\linewidth]{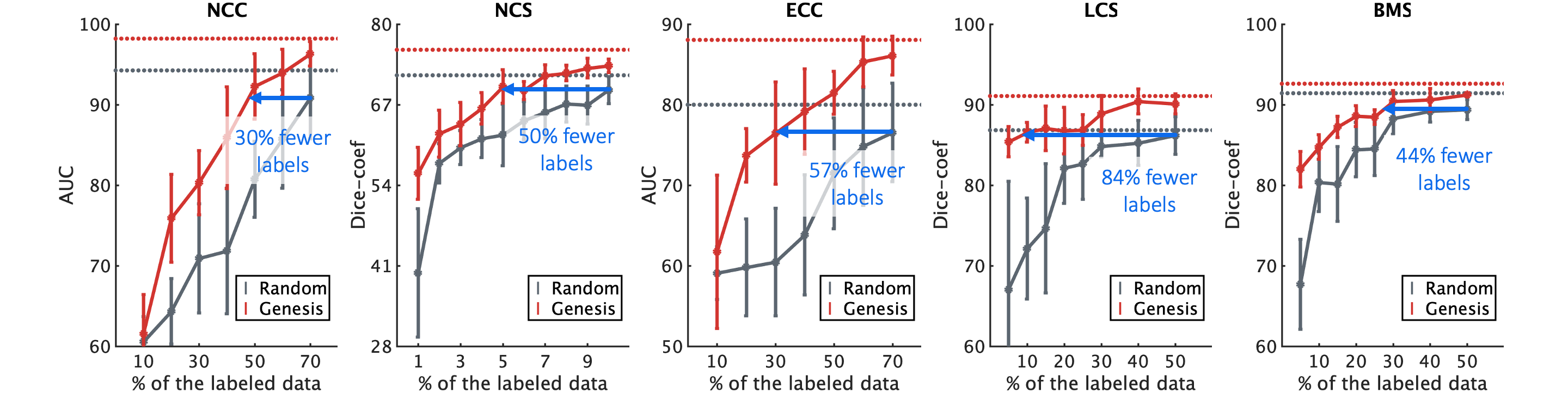}
\caption{
Initializing with our Models Genesis, the annotation cost can be reduced by 30$\%$, 50$\%$, 57$\%$, 84$\%$, and 44$\%$ for target tasks \texttt{NCC}, \texttt{NCS}, \texttt{ECC}, \texttt{LCS}, and \texttt{BMS}, respectively. With decreasing amounts of labeled data, Models Genesis (red) retain a much higher performance on all five target tasks, whereas learning from scratch (grey) fails to generalize. Note that the horizontal red and gray lines refer to the performances that can eventually be achieved by Models Genesis and learning from scratch, respectively, when using the entire dataset.
}
\label{fig:annotation_cost}
\end{figure*}

\begin{figure*}[t]
\centering
\includegraphics[width=1.0\linewidth]{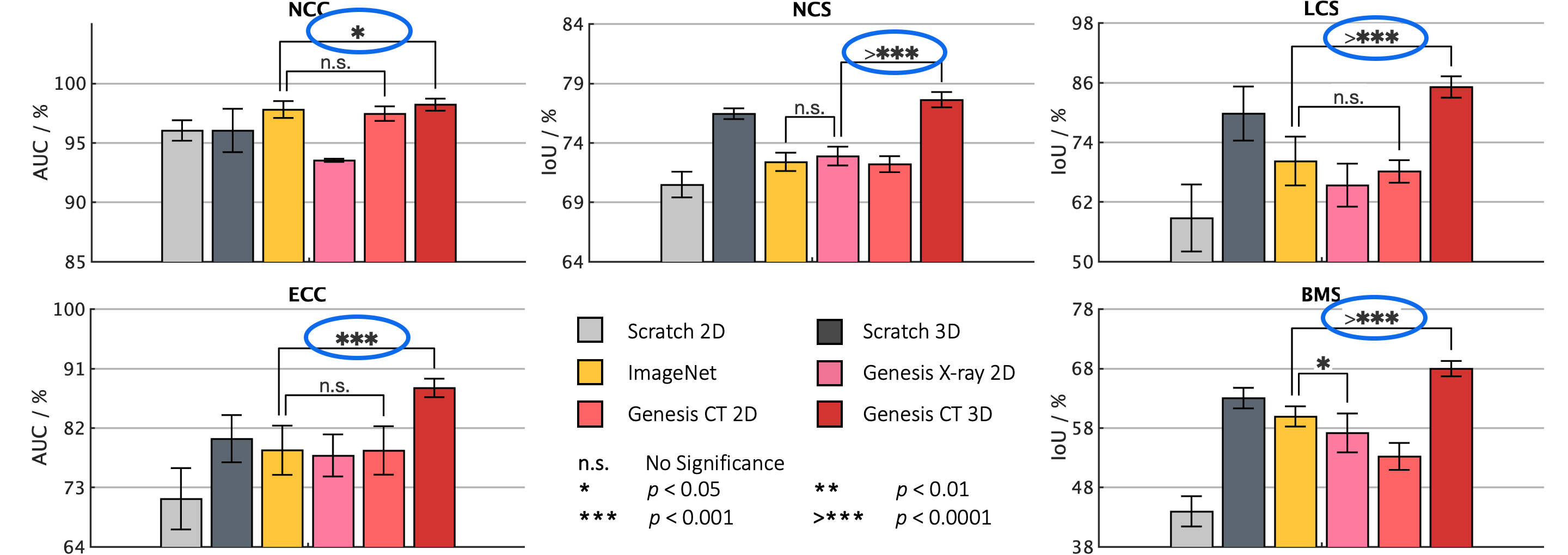}
\caption{
When solving problems in volumetric medical modalities, such as CT and MRI images, 3D \textit{volume-based} approaches consistently offer superior performance than 2D \textit{slice-based} approaches empowered by transfer learning. We conduct statistical analyses (circled in blue) between the highest performance achieved by 3D and 2D solutions. Training 3D models from scratch does not necessarily outperform their 2D counterparts (see \texttt{NCC}). However, training the same 3D models from Genesis Chest CT outperforms all their 2D counterparts, including fine-tuning Models ImageNet as well as fine-tuning our Genesis Chest X-ray and Genesis Chest CT 2D. It confirms the effectiveness of Genesis Chest CT in unlocking the power of 3D models. In addition, we have also provided statistical analyses between the highest and the second highest performances achieved by 2D models, finding that Models Genesis (2D) offer equivalent performances (n.s.) to Models ImageNet in four out of the five applications.
}
\label{fig:2D_3D_target_tasks}
\end{figure*}

\subsection{Models Genesis surpass existing pre-trained 3D models}
\label{sec:public_3d_model}

We have evaluated our Models Genesis with existing publicly available pre-trained 3D models on five distinct medical target tasks. As shown in \tablename~\ref{tab:top_existing_models}, Genesis Chest CT noticeably contrasts with any other existing 3D models, which have been pre-trained by full supervision. Note that, in the liver segmentation task (\texttt{LCS}), Genesis Chest CT is slightly outperformed by MedicalNet because of the benefit that MedicalNet gained from its (fully) supervised pre-training  on the LiTS dataset directly. Further statistical tests reveal that Genesis Chest CT still yields comparable performance with MedicalNet at $p=0.05$ level. For the rest four target tasks, Genesis Chest CT achieves superior performance against all its counterparts by a large margin, demonstrating the effectiveness and transferability of the learned features of Models Genesis, which are beneficial for both classification and segmentation tasks. 

More importantly, although Genesis Chest CT is pre-trained on Chest CT only, it can generalize to different organs, diseases, datasets, and even modalities. For instance, the target task of pulmonary embolism false positive reduction is performed in Contrast-Enhanced CT scans that can appear differently from the proxy tasks in normal CT scans; yet, Genesis Chest CT achieves a remarkable improvement over training from scratch, increasing the AUC by 7 points. Moreover, Genesis Chest CT continues to yield a significant IoU gain in liver segmentation even though the proxy task and target task are significantly different in both, diseases affecting the organs (lung vs.~liver) and the dataset itself (LUNA~2016 vs.~LiTS~2017). We have further examined Genesis Chest CT and other existing pre-trained models using MRI Flair images, which represent the widest domain distance between the proxy and target tasks. As reported in \tablename~\ref{tab:top_existing_models} (\texttt{BMS}), Genesis Chest CT yields nearly a 5-point improvement in comparison with random initialization. The increased performance on the MRI imaging task is a particularly strong demonstration of the transfer learning capabilities of our Genesis Chest CT. To further investigate the behavior of Genesis Chest CT when encountering medical images from different modalities, we have provided extensive visualization in \figurename~\ref{fig:xray_across_restoration}, including example images from CT, X-ray, Ultrasound, and MRI modalities.

Considering the model footprint, our Models Genesis take the basic 3D U-Net as the backbone, carrying much fewer parameters than the existing open-source pre-trained 3D models. For example, we have adopted MedicalNet with resnet-101 as the backbone, which offers the highest performance based on \citet{chen2019med3d} but comprises of 85.75M parameters; the pre-trained I3D~\citep{carreira2017quo} contains 25.35M parameters in the encoder; the pre-trained NiftyNet uses Dense V-Networks~\citep{gibson2018automatic} as backbone, comprising of only 2.60M parameters, but it does not perform as well as its counterparts in all five target tasks. Taken together, these results indicate that our Models Genesis, with only 16.32M parameters, surpass all existing pre-trained 3D models in terms of generalizability, transferability, and parameter efficiency.

\subsection{Models Genesis reduce annotation efforts by at least 30$\%$}
\label{sec:annotation_effort}

While critics often stress the need for sufficiently large amounts of labeled data to train a deep model, transfer learning leverages the knowledge about medical images already learned by pre-trained models and therefore requires considerably fewer annotated data and training iterations than learning from scratch. We have simulated the scenarios of using a handful of labeled data, which allows investigating the power of our Models Genesis in transfer learning. \figurename~\ref{fig:annotation_cost} displays the results of training with a partial dataset, demonstrating that fine-tuning Models Genesis saturates quickly on the target tasks since it can achieve similar performance compared with the full dataset training. 
Specifically, the performance of learning 3D models from scratch with entire datasets can be approximated using Models Genesis with only 50$\%$, 5$\%$, 30$\%$, 5$\%$, and 30$\%$ of datasets for \texttt{NCC}, \texttt{NCS}, \texttt{ECC}, \texttt{LCS}, and \texttt{BMS}, respectively. 
This shows that our Models Genesis can mitigate the lack of labeled images, resulting in a more annotation efficient deep learning in the end. 

Furthermore, the performance gap between fine-tuning and learning from scratch is significant and steady over training models with each partial data point. For the lung nodule false positive reduction target task (\texttt{NCC} in~\figurename~\ref{fig:annotation_cost}), using only 49$\%$ training data, Models Genesis equal the performance of 70$\%$ training data learning from scratch. Therefore, about 30$\%$ of the annotation cost associated with learning from scratch in \texttt{NCC} is recovered by initializing with Models Genesis. For the lung nodule segmentation target task (\texttt{NCS} in~\figurename~\ref{fig:annotation_cost}), with 5$\%$ training data, Models Genesis can achieve the performance equivalent to learning from scratch using 10$\%$ training data. Based on this analysis, the cost of annotation in \texttt{NCS} can be reduced by half using Models Genesis compared with learning from scratch. For the pulmonary embolism false positive reduction target task (\texttt{ECC}),~\figurename~\ref{fig:annotation_cost} suggests that with only 30$\%$ training samples, Models Genesis achieve performance equivalent to learning from scratch using 70$\%$ training samples. Therefore, nearly 57$\%$ of the labeling cost associated with the use of learning from scratch for \texttt{ECC} could be recovered with our Models Genesis. For the liver segmentation target task (\texttt{LCS}) in~\figurename~\ref{fig:annotation_cost}, using 8$\%$ training data, Models Genesis equal the performance of learning from scratch using 50$\%$ training samples. Therefore, about 84$\%$ of the annotation cost associated with learning from scratch in \texttt{LCS} is recovered by initializing with Models Genesis. For the brain tumor segmentation target task (\texttt{BMS}) in~\figurename~\ref{fig:annotation_cost}, with less than 28$\%$ training data, Models Genesis achieve the performance equivalent to learning from scratch using 50$\%$ training data. Therefore, nearly 44$\%$ annotation efforts can be reduced using Models Genesis compared with learning from scratch. Overall, at least 30$\%$ annotation efforts have been reduced by Models Genesis, in comparison with learning a 3D model from scratch in five target tasks. With such annotation-efficient 3D transfer learning paradigm, computer-aided diagnosis of rare diseases or rapid response to global pandemics, which are severely underrepresented owing to the difficulty of collecting a sizeable amount labeled data, could be eventually actualized.

\begin{table}[t]
\centering
\footnotesize
\caption{
Our 3D approach, initialized by Models Genesis, significantly elevates the classification performance compared with 2.5D and 2D approaches in reducing lung nodule and pulmonary embolism false positives. The entries in bold highlight the best results achieved by different approaches. For the 2D slice-based approach, we extract input consisting of three adjacent axial views of the lung nodule or pulmonary embolism and some of their surroundings. For the 2.5D orthogonal approach, each input is composed of an axial, coronal, and sagittal slice and centered at a lung nodule or pulmonary embolism candidate.
}
\label{tab:3d_2.5d_2d}
\begin{tabular}{p{0.36\linewidth}P{0.15\linewidth}P{0.15\linewidth}P{0.15\linewidth}}
    \shline
    Task: \texttt{NCC} & Random & ImageNet & Genesis \\
    \hline
    2D slice-based input & 96.03$\pm$0.86 & \textbf{97.79$\pm$0.71} & 97.45$\pm$0.61 \\
    2.5D orthogonal input & 95.76$\pm$1.05 & \textbf{97.24$\pm$1.01} & 97.07$\pm$0.92 \\
    3D volume-based input & 96.03$\pm$1.82 & n/a & \textbf{98.34$\pm$0.44} \\
    \hline
    \hline
    Task: \texttt{ECC} & Random & ImageNet & Genesis \\
    \hline
    2D slice-based input & 60.33$\pm$8.61 & 62.57$\pm$8.04 & \textbf{62.84$\pm$8.78} \\
    2.5D orthogonal input & 71.27$\pm$4.64 & \textbf{78.61$\pm$3.73} & 78.58$\pm$3.67 \\
    3D volume-based input & 80.36$\pm$3.58 & n/a & \textbf{88.04$\pm$1.40} \\
    \shline
    \end{tabular}
\end{table}

\subsection{Models Genesis consistently top any 2D/2.5D approaches}
\label{sec:models_genesis_top_2D}

We have thus far presented the power of 3D models in processing volumetric data, in particular, with limited annotation. Besides adopting 3D models, another common strategy to handle limited data in volumetric medical imaging is to reformat 3D data into a 2D image representation followed by fine-tuning pre-trained Models ImageNet~\citep{shin2016deep,tajbakhsh2016convolutional}. This approach increases the training examples by order of magnitude, but it sacrifices the 3D context. It is interesting to note how Genesis Chest CT compares with this \textit{de facto} standard in 2D. We have thus implemented two different methods to reformat 3D data into 2D input: the regular 2D representation obtained by extracting adjacent axial slices~\citep{ben2016fully,sun2017multiphase}, and the 2.5D representation~\citep{prasoon2013deep,roth2014new,roth2015improving} composed of axial, coronal, and sagittal slices from volumetric data. Both of these 2D approaches seek to use 2D representation to emulate something three dimensional, in order to fit the paradigm of fine-tuning Models ImageNet. 
In the inference, classification and segmentation tasks are evaluated differently in 2D: for classification, the model predicts labels of slices extracted from the center locations because other slices are not guaranteed to include objects; for segmentation, the model predicts segmentation mask slice by slice and form the 3D segmentation volume by simply stacking the 2D segmentation maps.

\figurename~\ref{fig:2D_3D_target_tasks} exposes the comparison between 3D and 2D models on five 3D target tasks. Additionally, \tablename~\ref{tab:3d_2.5d_2d} compares 2D slice-based, 2.5D orthogonal, and 3D volume-based approaches on lung nodule and pulmonary embolism false positive reduction tasks. As evidenced by our statistical analyses, the 3D models trained from Genesis Chest CT achieve significantly higher average performance and lower standard deviation than 2D models fine-tuned from ImageNet using either 2D or 2.5D image representation. Nonetheless, the same conclusion does not apply to the models trained from scratch---3D scratch models are outperformed by 2D models in one out of the five target tasks (\ie \texttt{NCC} in \figurename~\ref{fig:2D_3D_target_tasks} and \tablename~\ref{tab:3d_2.5d_2d}) and also exhibit an undesirably larger standard deviation. We attribute the mixed results of 3D scratch models to the larger number of model parameters and limited sample size in the target tasks, which together impede the full utilization of 3D context. In fact, the undesirable performance of the 3D scratch models highlights the effectiveness of Genesis Chest CT, which unlocks the power of 3D models for medical imaging. To summarize, we believe that 3D problems in medical imaging should be solved in 3D directly.

\section{Discussions}
\label{sec:discussion}

\subsection{Do we still need a medical ImageNet?}
\label{sec:medical_imagenet}

In computer vision, at the time this paper is written, no self-supervised learning method outperforms fine-tuning models pre-trained on ImageNet~\citep{jing2020self,chen2019self,kolesnikov2019revisiting,zhou2019models,hendrycks2019using,zhang2019aet,caron2019unsupervised}. Therefore, it may seem surprising to observe from our results in \tablename~\ref{tab:top_existing_models} that (fully) supervised representation learning methods do not necessarily offer higher performances in some 3D target tasks than self-supervised representation learning methods. We ascribe this phenomenon to the limited amount of supervision used in their pre-training (90 cases for NiftyNet~\citep{gibson2018niftynet} and 1,638 cases for MedicalNet~\citep{chen2019med3d}) or the domain distance (from videos to CT/MRI for I3D~\citep{carreira2017quo}). Evidenced by a prior study~\citep{sun2017revisiting} on ImageNet pre-training, large amount of supervision is required to foster a generic, comprehensive image representation. Back in 2009, when ImageNet had not been established, it was challenging to empower a deep model with generic image representation using a small or even medium size of labeled data, the same situation, we believe, that presents in 3D medical image analysis today. Therefore, despite the outstanding performance of Models Genesis, there is no doubt that a large, strongly annotated dataset for medical image analysis, like ImageNet~\citep{deng2009imagenet} for computer vision, is still highly demanded. One of our goals for developing Models Genesis is to help create such a medical ImageNet. Based on a small set of expert annotations, models fine-tuned from Models Genesis will be able to help quickly generate initial rough annotations of unlabeled images for expert review, thus reducing the annotation efforts and accelerating the creation of a large, strongly annotated, medical ImageNet. In summary, Models Genesis are not designed to replace such a large, strongly annotated dataset for medical image analysis, as ImageNet for computer vision, but rather to help create one.

\subsection{Same-domain or cross-domain transfer learning?}
\label{sec:same_cross_modality}

Same-domain transfer learning is always preferred whenever possible because a relatively smaller domain gap makes the learned image representation more beneficial for target tasks. Even the most recent self-supervised learning approaches in medical imaging were solely evaluated within the same dataset, such as~\citet{chen2019self,tajbakhsh2019surrogate,zhu2020rubik}. Same-domain transfer learning strikes as a preferred choice in terms of performance; however, most of the existing medical datasets, with less than hundred cases, are usually too small for deep models to learn reliable image representation. Therefore, for our future work, we plan to combine the publicly available datasets from similar domains together to train modality-oriented models, including Genesis CT, Genesis MRI, Genesis X-ray, and Genesis Ultrasound, as well as organ-oriented models, including Genesis Brain, Genesis Lung, Genesis Heart, and Genesis Liver. 

Cross-domain transfer learning in medical imaging is the Holy Grail. Retrieving a large number of unlabeled images from a PACS system requires an IRB approval, often a long process; the retrieved images must be de-identified; organizing the de-identified images in a way suitable for deep learning is tedious and laborious. Therefore, large quantities of unlabeled datasets may not be readily available to many target domains. Evidenced by our results in \tablename~\ref{tab:top_existing_models} (\texttt{BMS}), Models Genesis have a great potential for cross-domain transfer learning; particularly, our distortion-based approaches (such as non-linear and local-shuffling) take advantage of relative intensity values (in all modalities) to learn shapes and appearances of various organs. Therefore, as our future work, we will be focusing on methods that generalize well across domains.

\subsection{Is any data augmentation suitable as a transformation?} 
\label{sec:choice_transformations}

We propose a self-supervised learning framework to learn image representation by discriminating and restoring images undergoing different transformations. One might argue that our image transformations can be interchangeable with existing data augmentation techniques~\citep{gan2015learning,wong2016understanding,perez2017effectiveness,shorten2019survey}, while we would like to make the distinction between these two concepts clearer. It is critical to assess whether a specific augmentation is practical and feasible for the image restoration task when designing image transformations. Simply introducing data augmentation can make a task ambiguous and lead to degenerate learning. To this end, we choose image transformations based on two principles:
\begin{itemize}
    \item First, the transformed sub-volume should not be found in the original CT scan. But it is possible to find a transformed sub-volume that has undergone such augmentations as rotation, flip, zoom in/out, or translation, as an alternative sub-volume in the original CT scan. In this scenario, without additional spatial information, the model would not be able to ``recover'' the original sub-volume by seeing the transformed one. As a result, we only elect the augmentations that can be applied to sub-volumes at the pixel level rather than the spatial level.
    \item Second, a transformation should be applicable for specific image properties. The augmentations that manipulate RGB channels, such as color shift and channel dropping, have little effect on CT/MRI images without the availability of color information. Instead, we promote brightness and contrast into monotonic color curves, resulting in a novel non-linear transformation, explicitly enabling the model to learn intensity distribution from medical images.
\end{itemize}
After filtering out using the above two principles, the remaining data augmentation techniques are not as many as expected. We have endeavored to produce learning perspective driven transformations rather than inviting any types of data augmentation into our framework. A recent study from \citet{chen2020simple} has also discovered a similar phenomenon: carefully designed augmentations are superior to autonomously discovered augmentations. This suggests a criterion of transformations driven by learning perspectives, in capturing a compelling, robust representation for 3D transfer learning in medical imaging.

\subsection{Can algorithms autonomously search for transformations?}
\label{sec:transformation_augmentation}

We follow two principles when designing suitable image transformations for our self-supervised learning framework (see Sec.~\ref{sec:choice_transformations}). Potentially, ``automated data augmentation'' can be considered as an efficient alternative because this line of research seeks to strip researchers from the burden of finding good parameterizations and compositions of transformations manually. Specifically, existing automated augmentation strategies reinforce models to learn an optimal set of augmentation policies by calculating the reward between predictions and image labels. To name a few, \citet{ratner2017learning} proposed a method for learning how to parameterize and composite the transformations for automated data augmentation, while preserving class \textit{labels} or null class for all data points. \citet{dao2019kernel} introduced a fast kernel alignment metric for augmentation selection. It requires image \textit{labels} for computing the kernel target alignment (as the reward) between the feature kernel and the label kernel. \citet{cubuk2019autoaugment} used reinforcement learning to form an algorithm that autonomously searches for preferred augmentation policies, magnitude, and probability for specific classification tasks, wherein the resultant accuracy of predictions and \textit{labels} is treated as the reward signal to train the recurrent network controller. \citet{wu2020generalization} proposed uncertainty-based sampling to select the most effective augmentation, but it is based on the highest loss that is computed between predictions and \textit{labels}. While the reward is well-defined in the aforementioned works, unfortunately, there is no available metric to determine the power of image representation directly; hence, no reward is readily established for representation learning. Rather than constrain the representation directly, our paper aims to design an image restoration task to let the model learn generic image representation from 3D medical images. To achieve this, inspired by~\citet{vincent2010stacked}, we modify the definition of a good representation into the following: ``\textit{a good representation is one that can be obtained robustly from a transformed input, and that will be useful for restoring the corresponding original input.}'' Consequently, mean square error (MSE) between the model's input and output is defined as the objective function in our framework. However, if we adopt MSE as the reward function, the existing automated augmentation strategies will end up selecting identical-mapping. This is because restoring images without any transformation is expected to give a lower error than restoring those with transformations. Evidenced by \figurename~\ref{fig:combined_vs_individuals}, identical-mapping results in a poor image representation. To summarize, the key challenge when employing automated augmentation strategies into our framework is how to define a proper reward for restoring images, and fundamentally, for learning image representation.

\begin{figure*}[t]
\begin{center}
\includegraphics[width=1.0\linewidth]{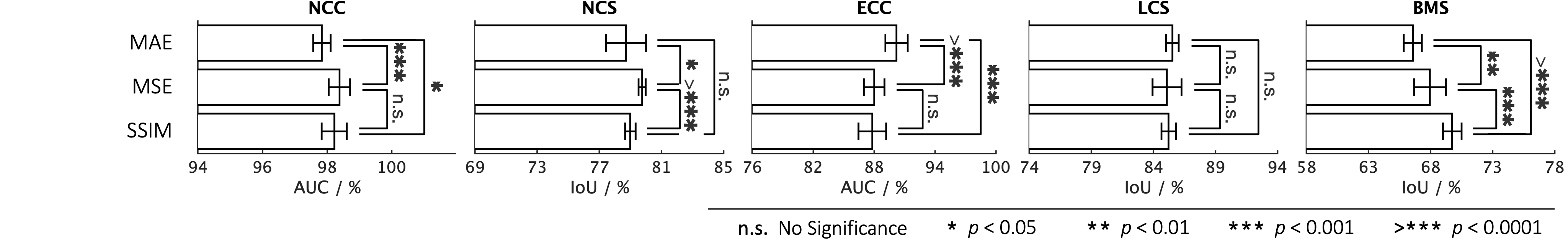}
\end{center}
\caption{
We compare three different losses for the task of image restoration. There is no evidence that the three losses have a decisive impact on the transfer learning results of five target tasks. Note that for this ablation study, all the proxy and target tasks are implemented in PyTorch.
}
\label{fig:losses}
\end{figure*}

\begin{figure}[t]
\centering
\includegraphics[width=1.0\linewidth]{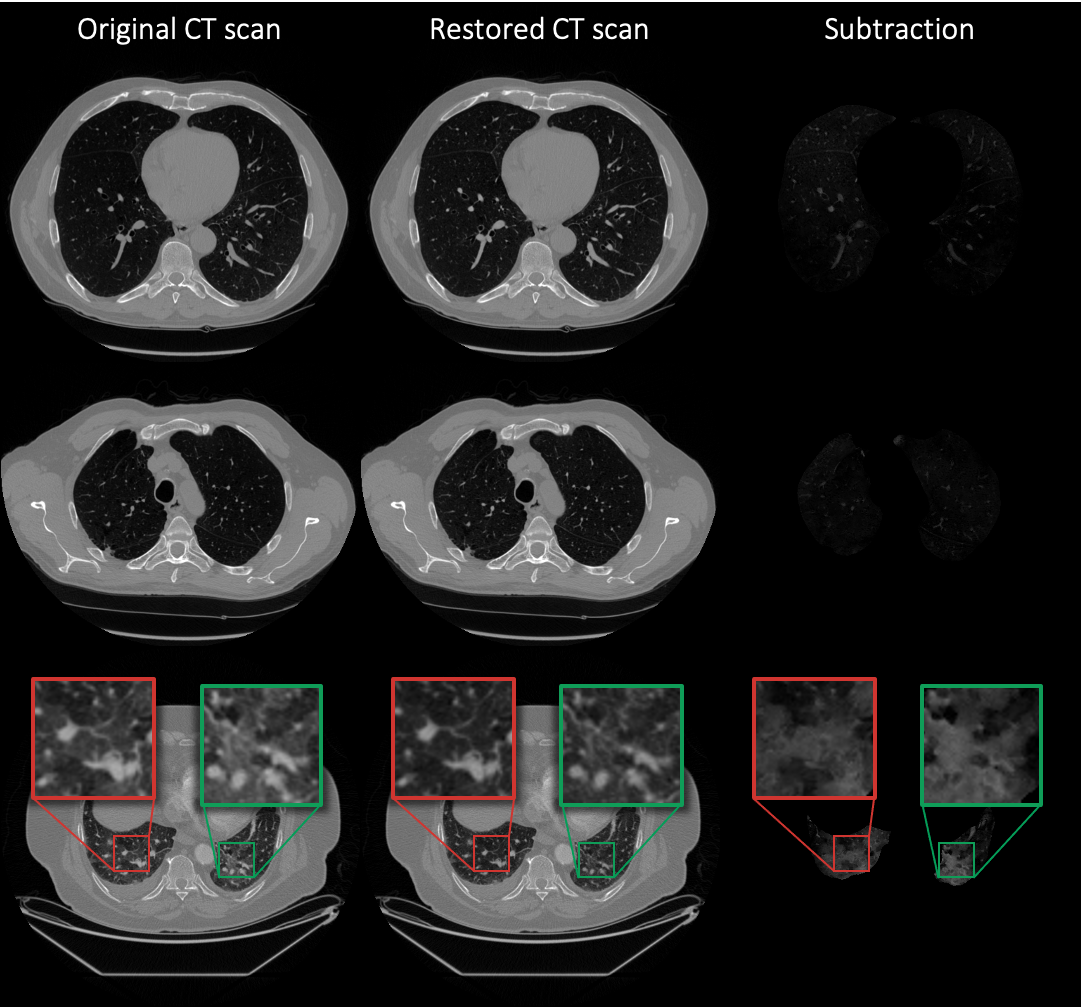}
\caption{
[Better viewed on-line and zoomed in for details] Examples of image restoration using Genesis Chest CT. We pass unseen CT images (Column 1) to the pre-trained model, obtaining the restored images (Column 2). The difference between input and output has been shown in Column 3. In most of the normal cases, such as those in Rows 1---2, Genesis Chest CT can perform a fairly reasonable identical-mapping. Meanwhile, for some cases that contain opacity in the lung, as illustrated in Row 3, Genesis Chest CT tends to restore a clearer lung. As a result, the diffuse region is revealed in the difference map automatically. We have zoomed in the region for a better visualization and comparison.
}
\label{fig:image_restoration}
\end{figure}

\subsection{How to assess restoration quality and its relationship to model transferability?}
\label{sec:restoration_loss}

Our transfer learning results in Sec.~\ref{sec:results} suggest that image restoration is a promising task to learn generic 3D image representation. This also means that image restoration quality has an implicit correlation with model transferability to some extent. To assess restoration quality, we compare the Mean Square Error (MSE) loss with other commonly used loss functions for image restoration, such as Mean Absolute Error (MAE) and Structural Similarity Index (SSIM)~\citep{wang2004image}. All of them compute the distance between input and output images, while SSIM concentrates more on the restoration quality in terms of structural similarity than MSE and MAE. Since the publicly available 3D SSIM loss was implemented in PyTorch\footnote{\label{foot:ssim}SSIM loss in 3D: \href{https://github.com/jinh0park/pytorch-ssim-3D}{github.com/jinh0park/pytorch-ssim-3D}}, to make the comparisons fair, we have adapted our five target tasks into PyTorch as well. 
\figurename~\ref{fig:losses} shows mixed performances of the five target tasks among the three alternative loss functions. As discussed in Sec.~\ref{sec:transformation_augmentation}, the ideal loss function for representation learning is one that can explicitly determine the power of image representation. However, the three losses explored in this section are implicit, based on the premise that the image restoration quality can indicate a good representation. 
Further studies with restoration quality assessment and its relationship to model transferability are therefore suggested.

\subsection{Could Models Genesis detect infected regions from images autonomously?}
\label{sec:disease_detection}

As referenced from Sec.~\ref{sec:experiments_pretraining_models_genesis}, Genesis Chest CT has been pre-trained using 623 CT images in the LUNA~2016 dataset. To assess the image restoration quality, we utilize the rest of the 265 CT images from the dataset and present examples in \figurename~\ref{fig:image_restoration}. Specifically, we pass the original CT images to the pre-trained Genesis Chest CT. To visualize the modifications, we have further plotted the difference maps by subtracting the input and output. Since the input images involve no image transformation, most of the restored CT scans (see Rows~1---2) can preserve the texture and structures of the input images, only encountering few changes thanks to the identical-mapping training scheme and the skip connections between encoder and decoder. Nonetheless, we observe some failed cases (see Row~3), especially when the input CT image contains diffuse disease, which appears as an opacity in the lung.
Genesis Chest CT happens to ``remove'' those opaque regions and restore a much clearer lung. This may be due to the fact that the majority of cropped sub-volumes are normal and are being used as ground truth, which empowers the pre-trained model with capabilities of detecting and restoring ``novelties'' in the CT scans. More specifically, in our work, these novelties include abnormal intensity distribution injected by non-linear transformation, atypical texture and boundary injected by local-shuffling, and discontinuity injected by both inner and outer cutout. Based on the surrounding anatomical structure, the model predicts the opaque area to be air, therefore restoring darker intensity values. This behavior is certainly a ``mistake'' in terms of image restoration, but it can also be thought of as an attempt to detect diffuse diseases in the lung, which is challenging to annotate due to their unclear boundary. By training an image restoration task, the diseased area will be revealed by simple \textit{subtraction} of the input and output. More importantly, this suggested detection approach requires zero human annotation, neither image-level label nor pixel-level contour, contrasting from the existing weakly supervised disease detection approaches~\citep{zhou2016learning,baumgartner2018visual,cai2018iterative,siddiquee2019learning}.

\section{Related Work}
\label{sec:related_works}


With the splendid success of deep neural networks, transfer learning~\citep{pan2010survey,weiss2016survey,yosinski2014transferable} has become integral to many applications, especially medical imaging~\citep{greenspan2016guest,litjens2017survey,lu2017deep,shen2017deep,wang2017comparison,zhou2017fine,zhou2019models}. This immense popularity of transfer learning is attributed to the learned image representation, which offers convergence speedups and performance gains for most target tasks, in particular, with limited annotated data. In the following sections, we review the works related to supervised and self-supervised representation learning.

\subsection{Supervised representation learning} 


ImageNet contains more than fourteen million images that have been manually annotated to indicate which objects are present in each image; and more than one million of the images have actually been annotated with the bounding boxes of the objects in the image. Pre-training a model on ImageNet and then fine-tuning it on different medical imaging tasks has seen the most practical adoption in medical image analysis~\citep{shin2016deep,tajbakhsh2016convolutional}. To classify the common thoracic diseases from ChestX-ray14 dataset, as evidenced in~\citet{irvin2019chexpert}, nearly all the leading methods~\citep{guan2018multi,guendel2018learning,ma2019multi,tang2018attention} follow the paradigm of ``fine-tuning Models ImageNet'' by adopting different architectures, such as ResNet~\citep{he2016deep} and DenseNet~\citep{huang2017densely}, along with their pre-trained weights. Other representative medical applications include identifying skin cancer from dermatologist level photographs~\citep{esteva2017dermatologist}, offering early detection of Alzheimer's Disease~\citep{ding2018deep}, and performing effective detection of pulmonary embolism~\citep{tajbakhsh2019computer}.

Despite the remarkable transferability of Models ImageNet, pre-trained 2D models offer little benefits towards 3D medical imaging tasks in the most prominent medical modalities (\eg CT and MRI). To fit this paradigm, 3D imaging tasks have to be reformulated and solved in 2D or 2.5D~\citep{roth2015improving,roth2014new,tajbakhsh2015computer}, thus losing rich 3D anatomical information and inevitably compromising the performance. Annotating 3D medical images at the similar scale with ImageNet requires a significant research effort and budget. It is currently not feasible to create annotated datasets comparable to this size for every 3D medical application. 
Consequently, for lung cancer risk malignancy estimation, \citet{ardila2019end} resorted to incorporate 3D spatial information by using Inflated 3D (I3D)~\citep{carreira2017quo}, trained from the Kinetics dataset, as the feature extractor. Evidenced by \tablename~\ref{tab:top_existing_models}, it is not the most favorable choice owing to the large domain gap between the temporal video and medical volume. This limitation has led to the development of model zoo in NiftyNet~\citep{gibson2018niftynet}. However, they were trained with small datasets for specific applications (\eg brain parcellation and organ segmentation), and were never intended as source models for transfer learning. Our experimental results in \tablename~\ref{tab:top_existing_models} indicate that NiftyNet models offer limited benefits to the five target medical applications via transfer learning. More recently, \citet{chen2019med3d} have pre-trained 3D residual network by jointly segmenting the objects annotated in a collection of eight medical datasets, resulting in MedicalNet for 3D transfer learning. In \tablename~\ref{tab:top_existing_models}, we have examined the pre-trained MedicalNet on five target tasks in comparison with our Models Genesis. As reviewed, each and every aforementioned pre-trained model requires massive, high-quality annotated datasets. However, seldom do we have a perfectly-sized and systematically-labeled dataset to pre-train a deep model in medical imaging, where both data and annotations are expensive to acquire. We overcome the above limitation via self-supervised learning, which allows models to learn image representation from abundant unlabeled medical image data with {\em zero} human annotation effort.

\subsection{Self-supervised representation learning}

Aiming at learning image representation from unlabeled data, self-supervised learning research has recently experienced a surge in computer vision~\citep{caron2018deep,chen2019rotation,doersch2015unsupervised,goyal2019scaling,jing2020self,mahendran2018cross,mundhenk2018improvements,noroozi2018boosting,noroozi2016unsupervised,pathak2016context,sayed2018cross,zhang2016colorful,zhang2017split}, but it is a relatively new trend in modern medical imaging. The key challenge for self-supervised learning is identifying a suitable self supervision task, \ie generating input and output instance pairs from the data. Two of the preliminary studies include predicting the distance and 3D coordinates of two patches randomly sampled from the same brain~\citep{spitzer2018improving}, identifying whether two scans belong to the same person, and predicting the level of vertebral bodies~\citep{jamaludin2017self}. Nevertheless, these two works are incapable of learning representation from ``self-supervision'' because they demand auxiliary information and specialized data collection such as paired and registered images. By utilizing only the original pixel/voxel information shipped with data, several self-supervised learning schemes have been developed for different medical applications: \citet{ross2018exploiting} adopted colorization as proxy task, wherein color colonoscopy images are converted to gray-scale and then recovered using a conditional Generative Adversarial Network (GAN); \citet{alex2017semisupervised} pre-trained a stack of denoising auto-encoders, wherein the self-supervision was created by mapping the patches with the injected noise to the original patches; \citet{chen2019self} designed image restoration as proxy task, wherein small regions were shuffled within images and then let models learn to restore the original ones; \citet{zhuang2019self} and \citet{zhu2020rubik} introduced a 3D representation learning proxy task by recovering the rearranged and rotated Rubik's cube; and finally \citet{tajbakhsh2019surrogate} individualized self-supervised schemes for a set of target tasks. As seen, the previously discussed self-supervised learning schemes, both in computer vision and medical imaging, are developed individually for specific target tasks, therefore, the generalizability and robustness of the learned image representation have yet to be examined across multiple target tasks. To our knowledge, we are the first to investigate cross-domain self-supervised learning in medical imaging. 

\subsection{Our previous work}
\label{sec:previous_work}

\citet{zhou2019models} first presented generic autodidactic models for 3D medical imaging, which obtain common image representation that is transferable and generalizable across diseases, organs and modalities, overcoming the scalablity issue associated with multiple tasks. This paper extends the preliminary version substantially with the following improvements.
\begin{enumerate}
    \item We have introduced notations, formulas, and diagrams, as well as detailed methodology descriptions along with their learning objectives, for a succinct framework overview in Sec.~\ref{sec:method}.
    \item We have extended the brain tumor segmentation experiment using MRI Flair images in Sec.~\ref{sec:finetuning_models_genesis}, highlighting the transfer learning capabilities of Models Genesis from CT to MRI Flair domains.
    \item We have conducted comprehensive ablation studies between the combined scheme and each of the individual learning schemes in Sec.~\ref{sec:individual_combination}, demonstrating that learning from multiple perspectives leads to a more robust target task performance.
    \item We have investigated three different random initialization methods for 3D models in Sec.~\ref{sec:surpass_scratch}, suggesting that initializing with Models Genesis can offer much higher performances and faster convergences.
    \item We have examined Models Genesis with the existing pre-trained 3D models on five distinct medical target tasks in Sec.~\ref{sec:public_3d_model}, showing that with fewer parameters, Models Genesis surpass all publicly available 3D models in both generalizability and transferability.
    \item We have provided experimental results on five target tasks using limited annotated data in Sec.~\ref{sec:annotation_effort}, indicating that transfer learning from our Models Genesis can reduce annotation efforts by at least 30\%.
    \item We have investigated 3D sub-volume based approaches compared with 2D approaches fine-tuning from Models ImageNet using 2D/2.5D representation, underlining the power of pre-trained 3D models in Sec.~\ref{sec:models_genesis_top_2D}.
    
\end{enumerate}

\section{Conclusion}
\label{sec:conclusion}

A key contribution of ours is a collection of \textit{generic source} models, nicknamed Models Genesis, built directly from {\em unlabeled} 3D imaging data with our novel unified self-supervised method, for generating powerful application-specific \textit{target} models through transfer learning. While the empirical results are strong, surpassing state-of-the-art performances in most of the applications, our goal is to extend our Models Genesis to modality-oriented models, such as Genesis MRI and Genesis Ultrasound, as well as organ-oriented models, such as Genesis Brain and Genesis Heart. We envision that Models Genesis may serve as a primary source of transfer learning for 3D medical imaging applications, in particular, with limited annotated data. 
To benefit the research community, we make the development of Models Genesis open science, releasing our codes and models to the public. 
Creating all Models Genesis, an ambitious undertaking, takes a village; therefore, we would like to invite researchers around the world to contribute to this effort, and hope that our collective efforts will lead to the Holy Grail of Models Genesis, all powerful across diseases, organs, and modalities.

\section*{Acknowledgments}
This research has been supported partially by ASU and Mayo Clinic through a Seed Grant and an Innovation Grant, and partially by the National Institutes of Health (NIH) under Award Number R01HL128785. The content is solely the responsibility of the authors and does not necessarily represent the official views of the NIH.
This work has utilized the GPUs provided partially by the ASU Research Computing and partially by the Extreme Science and Engineering Discovery Environment (XSEDE) funded by the National Science Foundation (NSF) under grant number ACI-1548562.
We thank Z.~Guo for implementing Rubik's Cube~\citep{zhuang2019self} and the 3D version of Jigsaw~\citep{noroozi2016unsupervised} and DeepCluster~\citep{caron2018deep}; F.~Haghighi and M.~R.~Hosseinzadeh Taher for implementing the 3D version of in-painting~\citep{pathak2016context}, patch-shuffling~\citep{chen2019self}, and working with Z.~Guo in evaluating the performance of MedicalNet~\citep{chen2019med3d}; M.~M.~Rahman Siddiquee for examining NiftyNet~\citep{gibson2018niftynet} with our Models Genesis; P.~Zhang for comparing two additional random initialization methods with our Models Genesis; S.~Bajpai for comparing three loss functions of the proxy task; N.~Tajbakhsh for revising our conference paper; R.~Feng for valuable discussions; and S.~Tatapudi for helping improve the writing of this paper. The content of this paper is covered by patents pending.

\bibliographystyle{model2-names.bst}\biboptions{authoryear}
\bibliography{refs}

\newpage
\appendix

\section{Implementation details of revised baselines}
\label{sec:baseline_implementation_appendix}

\begin{figure*}[t]
\begin{center}
\includegraphics[width=1.0\linewidth]{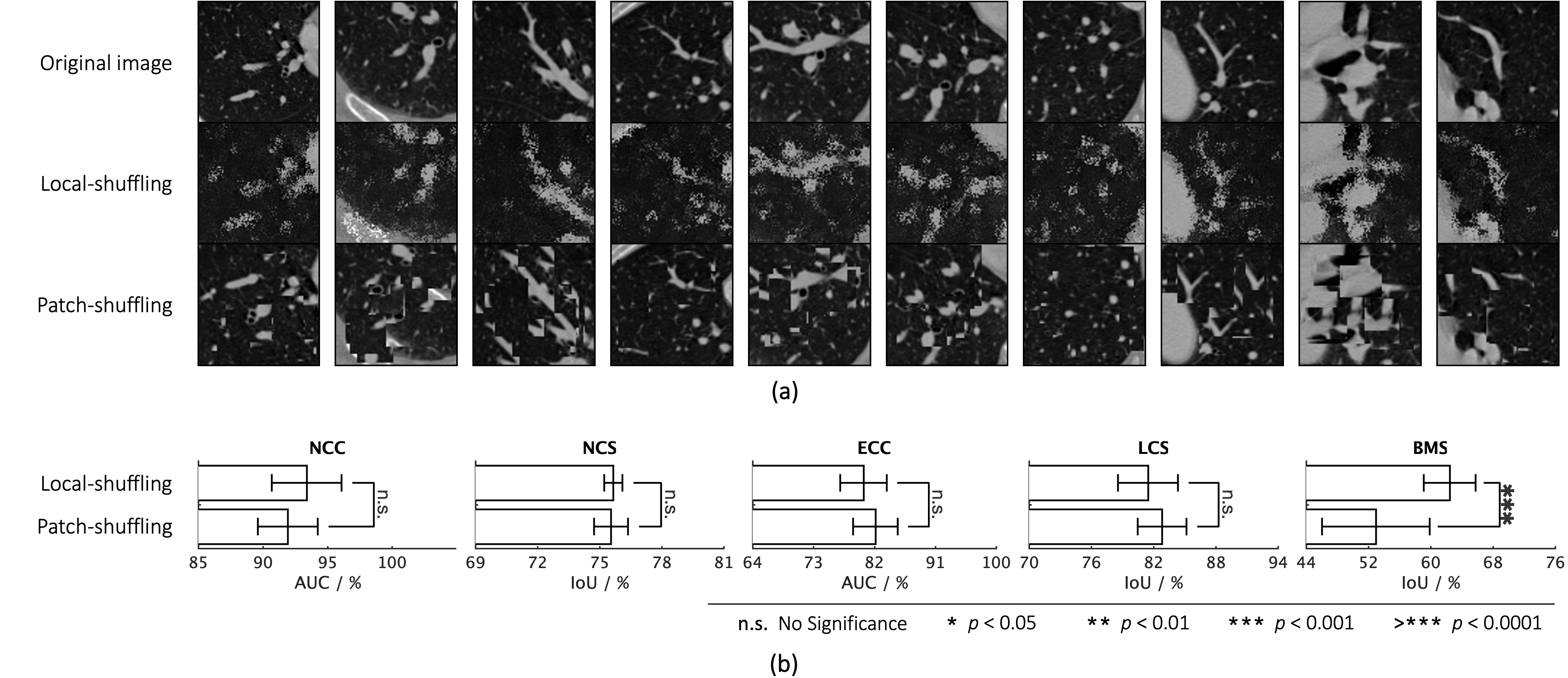}
\end{center}
\caption{
A direct comparison between global patch shuffling~\citep{chen2019self} and our local pixel shuffling. (a) illustrates ten example images undergone local-shuffling and patch-shuffling independently. As seen, the overall anatomical structure such as individual organs, blood vessels, lymph nodes, and other soft tissue structures are preserved in the transformed image through local-shuffling.
(b) presents the performance on five target tasks, showing that models pre-trained by our local-shuffling noticeably outperform those pre-trained by patch-shuffling for cross-domain transfer learning (\texttt{BMS}). 
}
\label{fig:localshuffling_patchshuffling}
\end{figure*}

\begin{figure*}[t]
\begin{center}
\includegraphics[width=1.0\linewidth]{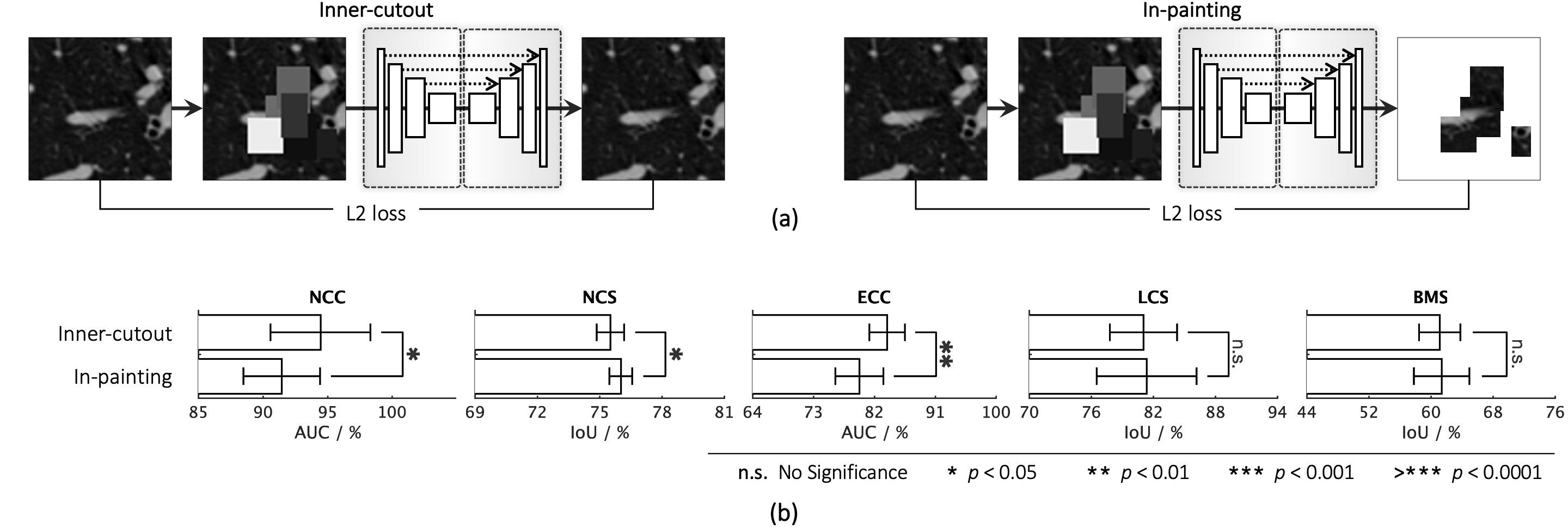}
\end{center}
\caption{
A direct comparison between image in-painting~\citep{pathak2016context} and our inner-cutout. (a) contrasts our inner-cutout with in-painting, wherein the model in the former scheme computes loss on the entire image and the model in the latter scheme computes loss only for the cutout area. (b) presents the performance on five target tasks, showing that inner-cutout is better suited for target classification tasks (\eg \texttt{NCC} and \texttt{ECC}), while in-painting is more helpful for target segmentation tasks (\eg \texttt{NCS}, \texttt{LCS}, and \texttt{BMS}).
}
\label{fig:innercutout_inpainting}
\end{figure*}

This work is among the first effort to create a comprehensive benchmark for existing self-supervised learning methods for 3D medical image analysis. We have extended the six most representative self-supervised learning methods into their 3D versions, including De-noising~\citep{vincent2010stacked}, In-painting~\citep{pathak2016context}, Jigsaw~\citep{noroozi2016unsupervised}, and Patch-shuffling~\citep{chen2019self}. These methods were originally introduced for the purpose of 2D imaging. 
On the other hand, the most recent 3D self-supervised method~\citep{zhuang2019self} learns representation by playing a Rubik's cube.
We have reimplemented it because their official implementation is not publicly available at the time this paper is written. All of the models are pre-trained using the LUNA~2016 dataset~\citep{setio2017validation} with the same sub-volumes extracted from CT scans as our models (see Sec.~\ref{sec:experiments_pretraining_models_genesis}). The detailed implementations of the baselines are elaborated in the following sections.

\subsection{Extended 3D De-noising} 
\label{sec:denoising_appendix}

In our 3D De-noising, which is inspired by its 2D counterpart~\citep{vincent2010stacked}, the model is trained to restore the original sub-volume from its transformed one with additive Gaussian noise (randomly sampling $\sigma\in[0,0.1]$). To correctly restore the original sub-volume, models are required to learn Gabor-like edge detectors when denoising transformed sub-volumes. Following the proposed image restoration training scheme, the auto-encoder network is replaced with a 3D U-Net, wherein the input is a $64\times 64\times 32$ sub-volume that has undergone Gaussian noise and the output is the restored sub-volume. The L2 distance between input and output is used as the loss function. 

\subsection{Extended 3D In-painting}
\label{sec:inpainting_appendix}

In our 3D In-painting, which is inspired by its 2D counterpart~\citep{pathak2016context}, the model is trained to in-paint arbitrary cutout regions based on the rest of the sub-volume. A qualitative illustration of the image in-painting task is shown in the right panel of \figurename~\ref{fig:innercutout_inpainting}(a). To correctly predict missing regions, networks are required to learn local continuities of organs in medical images via interpolation. Unlike the original in-painting, the adversarial loss and discriminator are excluded from our implementation of the 3D version because our primary goal is to empower models with generic representation, rather than generating sharper and realistic sub-volumes. The generator is a 3D U-Net, consisting of an encoder and a decoder. The input of the encoder is a $64\times 64\times 32$ sub-volume that needs to be in-painted. 
Their decoder works differently than our inner-cutout because it predicts the missing region only, and therefore, the loss is just computed on the cutout region---an ablation study on the loss has been further presented in \ref{sec:innercutout_inpainting}.

\subsection{Extended 3D Jigsaw}
\label{sec:jigsaw_appendix}

In our 3D Jigsaw, which is inspired by its 2D counterpart~\citep{noroozi2016unsupervised}, we utilize the implementation by~\citet{taleb20203d}\footnote{\label{foot:3d_self_learning}Self-Supervised 3D Tasks: \href{https://github.com/HealthML/self-supervised-3d-tasks}{github.com/HealthML/self-supervised-3d-tasks}}, wherein the puzzles are created by sampling a $3\times 3\times 3$ grid of 3D patches. Then, these patches are shuffled according to an arbitrary permutation, selected from a set of predefined permutations. This set with size $P=100$ is chosen out of the $(3\times 3\times 3)!$ possible permutations, by following the Hamming distance based algorithm, and each permutation is assigned an index. As a result, the problem is cast as a $P$-way classification task, \ie the model is trained to recognize the applied permutation index, allowing us to solve the 3D puzzles efficiently. We build the classification model by taking the encoder of 3D U-Net and appending a sequence of $fc$ layers. In the implementation, we minimize the cross-entropy loss of the list of extracted puzzles.

\subsection{Extended 3D Patch-shuffling}
\label{sec:patchshuffling_appendix}

In our 3D Patch-shuffling, which is inspired by its 2D counterpart~\citep{chen2019self}, the model learns image representation by restoring the image context. Given a sub-volume, we randomly select two isolated small 3D patches and swap their context. We set the length, width, and height of the 3D patch to be proportional to those in the entire sub-volume by 25\% to 50\%. Repeating this process for $T=10$ times can generate the transformed sub-volume (see examples in \figurename~\ref{fig:localshuffling_patchshuffling}(a)). The model is trained to restore the original sub-volume, where L2 distance between input and output is used as the loss function. To process volumetric input and ensure a fair comparison with other baselines, we replace their U-Net with 3D U-Net architecture, where the encoder and decoder serve as analysis and restoration parts, respectively.

\subsection{Extended 3D DeepCluster}
\label{sec:deepcluster_appendix}

In our 3D DeepCluster, which is inspired by its 2D counterpart~\citep{caron2018deep}, we iteratively cluster deep features extracted from sub-volumes by $k$-means and use the subsequent assignments as supervision to update the weights of the model. Through clustering, the model can obtain useful general-purpose visual features, requiring little domain knowledge and no specific signal from the inputs. We replaced original AlexNet/VGG architecture with the encoder of 3D U-Net to process 3D input sub-volumes. The number of clusters that works best for 2D tasks may not be a good choice for 3D tasks. To ensure a fair comparison, we extensively tune this hyper-parameter in $\{10,20,40,80,160,320\}$ and finally set to 260 from the narrowed down search space of $\{240,260,280\}$. Unlike ImageNet models for 2D imaging tasks, there is no available pre-trained 3D feature extractor for medical imaging tasks; therefore, we randomly initialize the model weights at the beginning. Our Models Genesis, the first generic 3D pre-trained models, could potentially be used as the 3D feature extractor and co-trained with 3D DeepCluster. 

\subsection{Rubik's cube}
\label{sec:rebikcube_appendix}

We implement Rubik's Cube with respect to~\citet{zhuang2019self}, which consists of cube rearrangement and cube rotation. Like playing a Rubik's cube, this proxy task enforces models to learn translational and rotational invariant features from raw 3D data. Given a sub-volume, we partition it into a $2\times 2\times 2$ grid of cubes. In addition to predicting orders (3D Jigsaw), this proxy task permutes the cubes with random rotations, forcing models to predict the orientation. Following the original paper, we limit the directions for cube rotation, \ie only allowing 180$^{\circ}$ horizontal and vertical rotations, to reduce the complexity of the task. The eight cubes are then fed into a Siamese network with eight branches sharing the same weight to extract features. The feature maps from the last fully-connected or convolution layer of all branches are concatenated and given as input to the fully-connected layer of separate tasks, \ie cube ordering and orienting, which are supervised by permutation loss and rotation loss, respectively, with equal weights.

\section{Configurations of publicly available models}
\label{sec:public_3d_model_appendix}

For publicly available models, we do not re-train their proxy tasks and instead simply endeavor to find the best hyper-parameters for each of them in target tasks. We compare them with our Models Genesis in a user perspective, which might seem to be unfair in a research perspective because many variables are asymmetric among the competitors, such as programming platform, model architecture, number of parameters, etc. However, the goal of this section is to experiment with existing ready-to-use pre-trained models under different medical tasks; therefore, we presume that all of the publicly available models and their configurations have been carefully composed to the optimal setting.

\subsection{NiftyNet}
\label{sec:niftynet_appendix}

We examine the effectiveness of fine-tuning from NiftyNet in five target tasks. We should note that NiftyNet is not initially designed for transfer learning but is one of the few publicly available supervised pre-trained 3D models. The model from~\citet{gibson2018automatic} has been considered as the baseline in our experiments because it has also been pre-trained on the chest region in CT modality and applied an encoder-decoder architecture that is similar to our work. We directly adopt the pre-trained weights of the dense V-Net architecture provided by NiftyNet, so it carries a smaller number of parameters than our 3D U-Net (2.60M vs. 16.32M). For target classification tasks, we use the dense V-Net encoder by appending a sequence of $fc$ layers; for target segmentation tasks, we use the entire dense V-Net. Since NiftyNet is developed in Tensorflow, all five target tasks are re-implemented using their build-in configuration. For each target task, we have tuned hyper-parameters (\eg learning rate and optimizer) and applied extensive data augmentations (\eg rotation and scaling).

\subsection{Inflated 3D}
\label{sec:i3d_appendix}

We download the Inflated 3D (I3D) model pre-trained from Flow streams in the Kinetics dataset~\citep{hara2018can} and fine-tune it on our five target tasks. The input sub-volume is copied into two channels to align with the required input shape. For target classification tasks, we take the pre-trained I3D and append a sequence of randomly initialized fully-connected layers. For target segmentation tasks, we take the pre-trained I3D as the encoder and expand a decoder to predict the segmentation map, resulting in a U-Net like architecture. The decoder is the same as that implemented in our 3D U-Net, consisting of up-sampling layers followed by a sequence of convolutional layers, batch normalization, and ReLU activation. Besides, four skip connections are built between the encoder and decoder, wherein feature maps before each pooling layer in the encoder are concatenated with same-scale feature maps in the decoder. All of the layers in the model are trainable during transfer learning. Adam method~\citep{kinga2015method} with a learning rate of $1e-4$ is used for optimization.

\subsection{MedicalNet}
\label{sec:medicalnet_appendix}

We download MedicalNet models~\citep{chen2019med3d} that have been pre-trained on eight publicly available 3D segmentation datasets. ResNet-50 and ResNet-101 backbones are chosen because they are reported by~\citet{chen2019med3d} as the most compelling backbones for target segmentation and classification tasks, respectively. Like I3D, we append a decoder at the end of the pre-trained encoder, randomly initialize its weights, and link the encoder with the decoder using skip connections. Owing to the 3D ResNet backbones, the resultant segmentation network for MedicalNet is much heavier than our 3D U-Net. To be consistent with the original programming platform of MedicalNet, we re-implement all five target tasks in PyTorch, using the same data separation and augmentation. We report the highest results achieved by any of the two backbones in \tablename~\ref{tab:top_existing_models}.

\section{Ablation experiments}
\label{sec:ablation_experiment_appendix}

\begin{figure}[t]
\begin{center}
\includegraphics[width=1.0\linewidth]{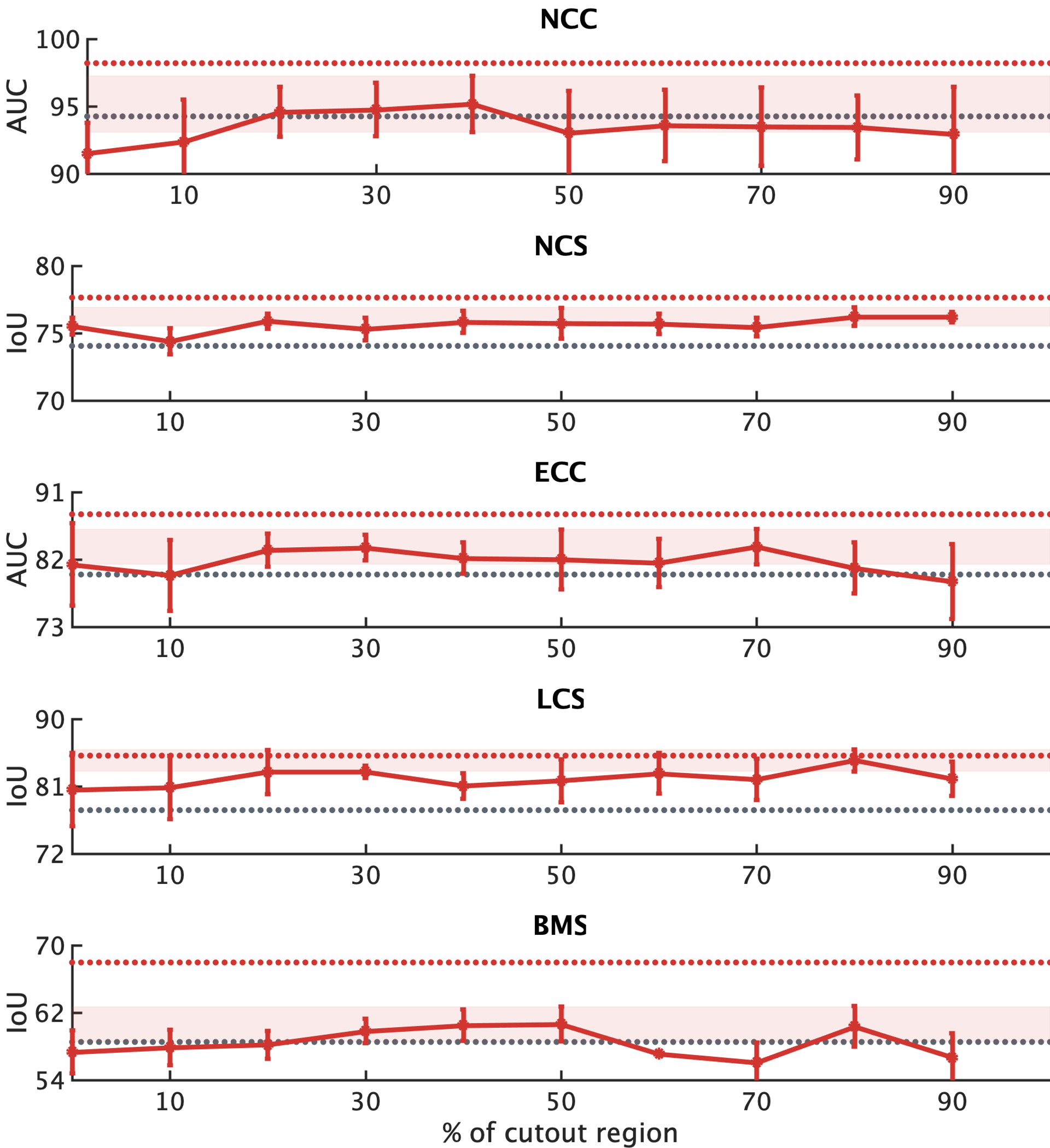}
\end{center}
\caption{
We extensively search for the optimal size of cutout regions spanning from 0\% to 90\%, incremented by 10\%. The points plotted within the red shade denote no significant difference ($p>0.05$) from the pinnacle from the curve. The horizontal red and gray lines refer to the performances achieved by Models Genesis and learning from scratch, respectively. This ablation study reveals that cutting 20\%---40\% of regions out could produce the most robust performance of target tasks. As a result, in our implementation, we cutout around 25\% regions from each sub-volume.
}
\label{fig:cutout_region_study}
\end{figure}

\subsection{Local pixel shuffling vs. global patch shuffling}
\label{sec:localshuffling_patchshuffling}

In the main paper, we have reported results of patch-shuffling~\citep{chen2019self} as a baseline in \tablename~\ref{tab:top_existing_models} and our local-shuffling in \figurename~\ref{fig:combined_vs_individuals}. To underline the value of preserving local and global structural consistency in the proxy task, we provide an explicit comparison between the two counterparts in \figurename~\ref{fig:localshuffling_patchshuffling}, arriving at three findings:
\begin{itemize}
    \item Global patch shuffling preserves local information while distorting global structure; local pixel shuffling maintains global structure but loses local details.
    \item For same-domain transfer learning (\eg pre-training and fine-tuning in CT images), global-shuffling and local-shuffling reveal no significant difference in terms of target task performance. Note that local-shuffling is preferable when recognizing small objects in target tasks (\eg pulmonary nodule and embolism), whereas patch-shuffling is beneficial for large objects (\eg brain tumor and liver).
    \item For cross-domain transfer learning (\eg pre-training in CT and fine-tuning in MRI images), models pre-trained by our local-shuffling noticeably outperform those pre-trained by patch-shuffling. 
\end{itemize}

\subsection{Compute loss on cutouts vs. entire images}
\label{sec:innercutout_inpainting}

The results of our ablation study for in-painting and inner-cutout on five target tasks are presented in~\figurename~\ref{fig:innercutout_inpainting}. We set all the hyper-parameters the same except for one factor: where to compute MSE loss, only cutout areas or the entire image. In general, there is a marginal difference in target segmentation tasks, but inner-cutout is superior to in-painting in target classification tasks. These results are in line with our hypothesis in Sec.~\ref{sec:experiments_pretraining_models_genesis}: the model must distinguish original versus transformed parts within the image, preserving the context if it is original and, otherwise, in-painting the context. Seemingly, in-painting that only computes loss on cutouts can fail to learn comprehensive representation as it is unable to leverage advancements from both ends. 

\subsection{Masked area size in outer-cutout}
\label{sec:cutout_region_study}

When applying cutout transformations to our self-supervised learning framework, we have one hyper-parameter to evaluate, \ie the size of cutout regions. Intuitively, it can influence the difficulty of the image restoration task. To explore the impact of this parameter on the performance of target tasks, we have conducted an ablation study to extensively search for the optimal value, spanning from 0\% to 90\%, incremented by intervals of 10\%. \figurename~\ref{fig:cutout_region_study} shows the performance of all five target tasks under different settings, suggesting that outer-cutout is robust to hyper-parameter changes to some extent. This finding is also consistent with that recommended in the original in-painting paper, where~\citet{pathak2016context} removed a number of smaller possibly overlapping masks, covering up to 1/4 of the image. Altogether, we finally cutout less than 1/4 of the entire sub-volume in both outer and inner cutout implementations.

\section{Qualitative assessment of image restoration}
\label{sec:qualitative_assessment}

\begin{figure*}[t]
\centering
\includegraphics[width=1.0\linewidth]{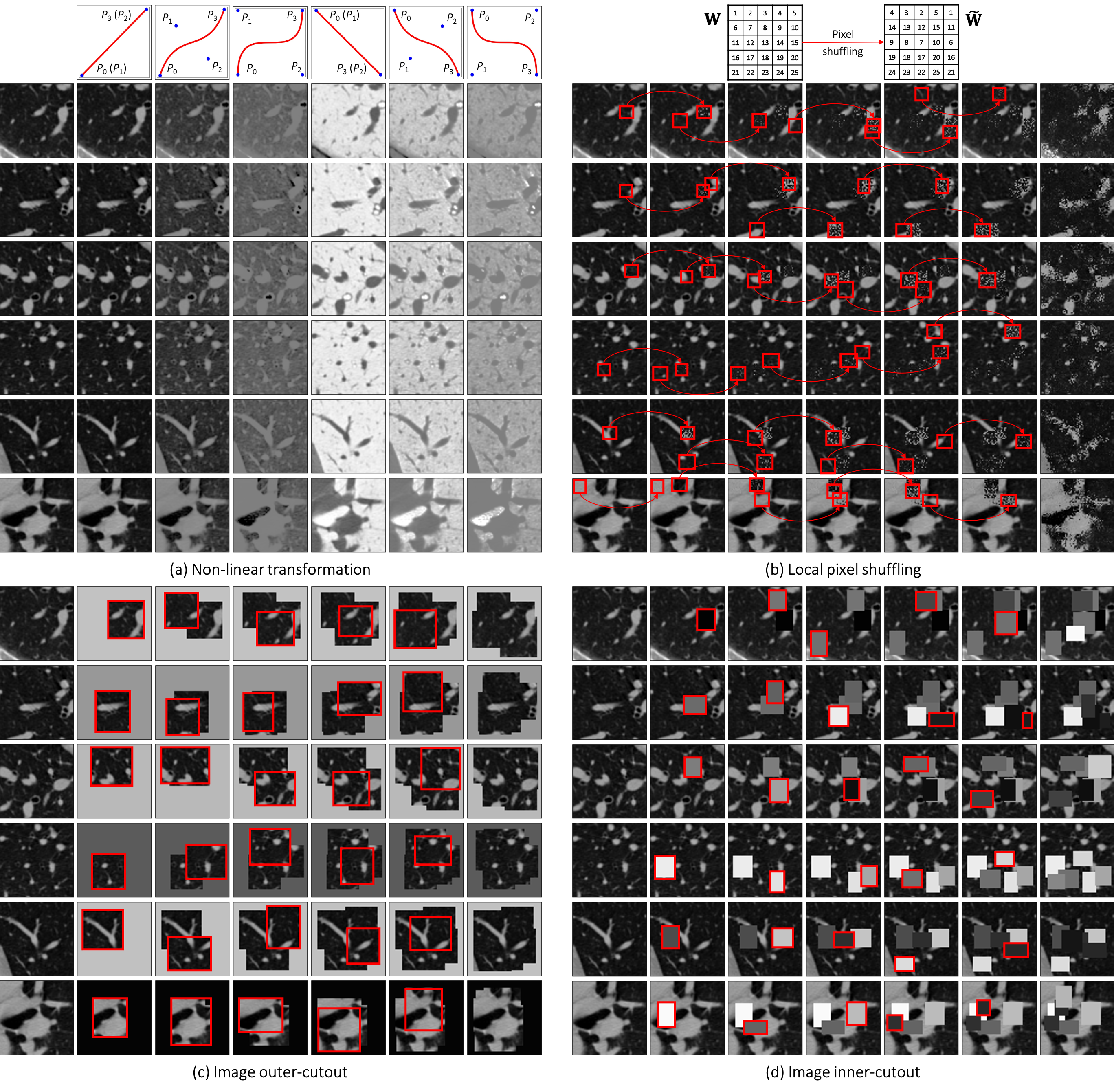}
\caption{
Illustration of the proposed four image transformations. For simplicity and clarity, we illustrate the transformation on a 2D CT slice, but our Genesis Chest CT is trained using 3D sub-volumes directly, transformed in a 3D manner; our 3D image transformations, with an exception of non-linear transformation, cannot be approximated in 2D. For ease of understanding, in (a) non-linear transformation, we have displayed an image undergoing different translating functions in Columns 2---7; in (b) local-shuffling, (c) outer-cutout, and (d) inner-cutout transformation, we have illustrated each of the processes step by step in Columns 2---6, where the first and last columns denote the original images and the final transformed images, respectively. In local-shuffling, a different window (b) is automatically generated and used in each step.
}
\label{fig:transformation_appendix}
\end{figure*}

\begin{figure*}[t]
\centering
\includegraphics[width=1.0\linewidth]{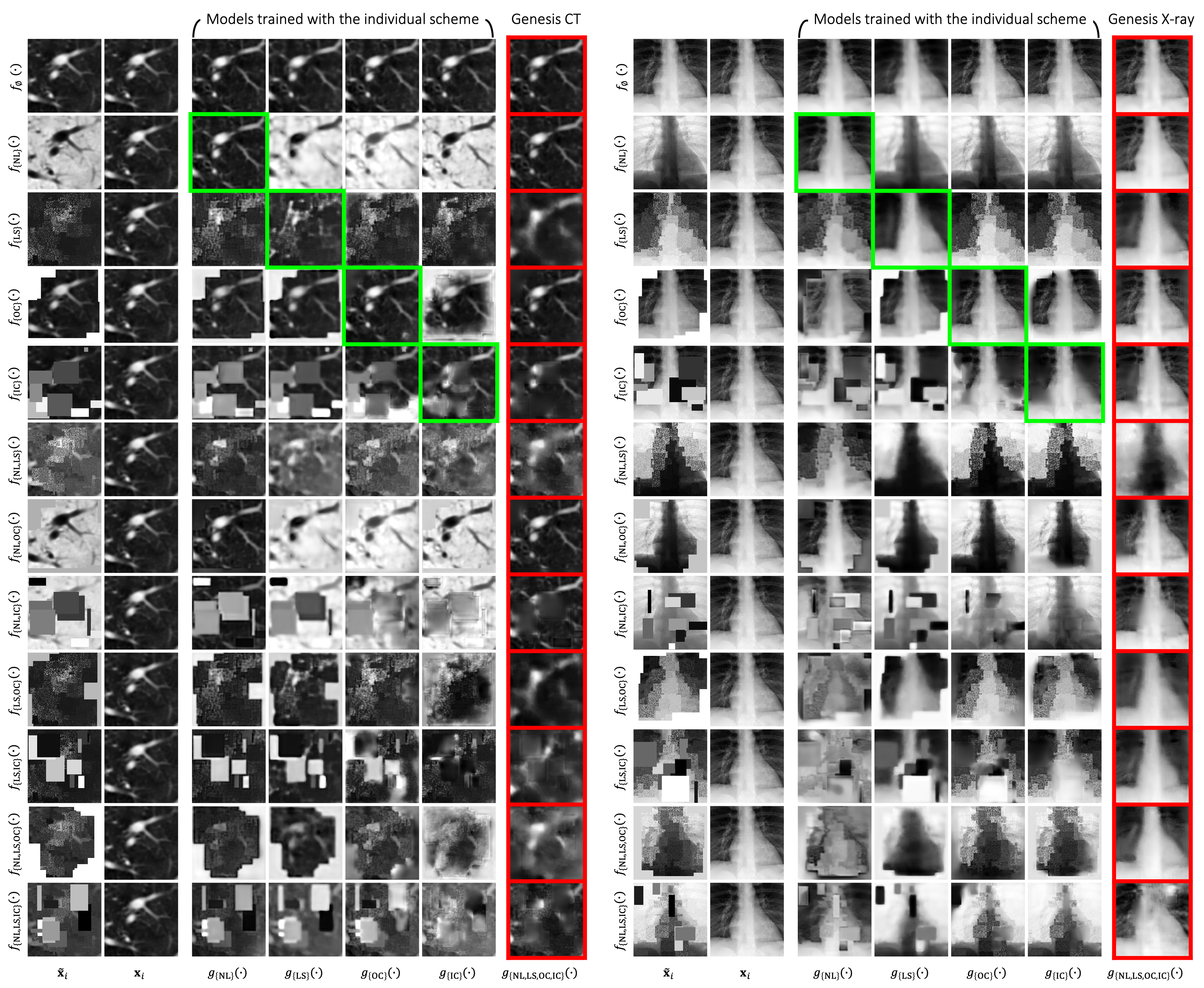}
\caption{
The left and right panels show the qualitative assessment of image restoration quality using Genesis CT and Genesis X-ray, respectively. These models are trained with different training schemes, including four individual schemes (Columns 3---6) and a combined scheme (Column 7). As discussed in \figurename~\ref{fig:self_supervised_learning_framework}, each original image $\textbf{x}_i$ can possibly undergo twelve different transformations. We test the models with all these possible twelve transformed images $\tilde{\textbf{x}}_i$. We specify types of the image transformation $f(\cdot)$ for each row and the training scheme $g(\cdot)$ for each column.
First of all, it can be seen that the models trained with individual schemes can restore previously unseen images that have undergone the same transformation very well (framed in green), but fail to handle other transformations. Taking non-linear transformation $f_{\text{(NL)}}(\cdot)$ as an example, any individual training scheme besides non-linear transformation itself cannot invert the pixel intensity from transformed whitish to the original blackish. As expected, the model trained with the combined scheme successfully restores original images from various transformations (framed in red). Second, the model trained with the combined scheme shows it is superior to other models even if they are trained with and tested on the same transformation. For example, in the local-shuffling case $f_{\text{(LS)}}(\cdot)$, the image recovered from the local-shuffling pre-trained model $g_{\text{(LS)}}(\cdot)$ is noisy and lacks texture. However, the model trained with the combined scheme $g_{\text{(NL,LS,OC,IC)}}(\cdot)$ generates an image with more underlying structures, which demonstrates that learning with augmented tasks can even improve the performance on each of the individual tasks. Third, the model trained with the combined scheme significantly outperforms models trained with individual training schemes when restoring images that have undergone seven different combined transformations (Rows 6---12). For example, the model trained with non-linear transformation $g_{\text{(NL)}}(\cdot)$ can only recover the intensity distribution in the transformed image undergone $f_{\text{(NL,IC)}}(\cdot)$ but leaves the inner cutouts unchanged.
These observations suggest that the model trained with the proposed unified self-supervised learning framework can successfully learn general anatomical structures and yield promising transferability on different target tasks. The quality assessment of image restoration further confirms our experimental observation, provided in Sec.~\ref{sec:individual_combination}, that the combined learning scheme exceeds each individual in transfer learning.
}
\label{fig:ct_restoration_individual_combined}
\end{figure*}

\begin{figure*}[t]
\centering
\includegraphics[width=1.0\linewidth]{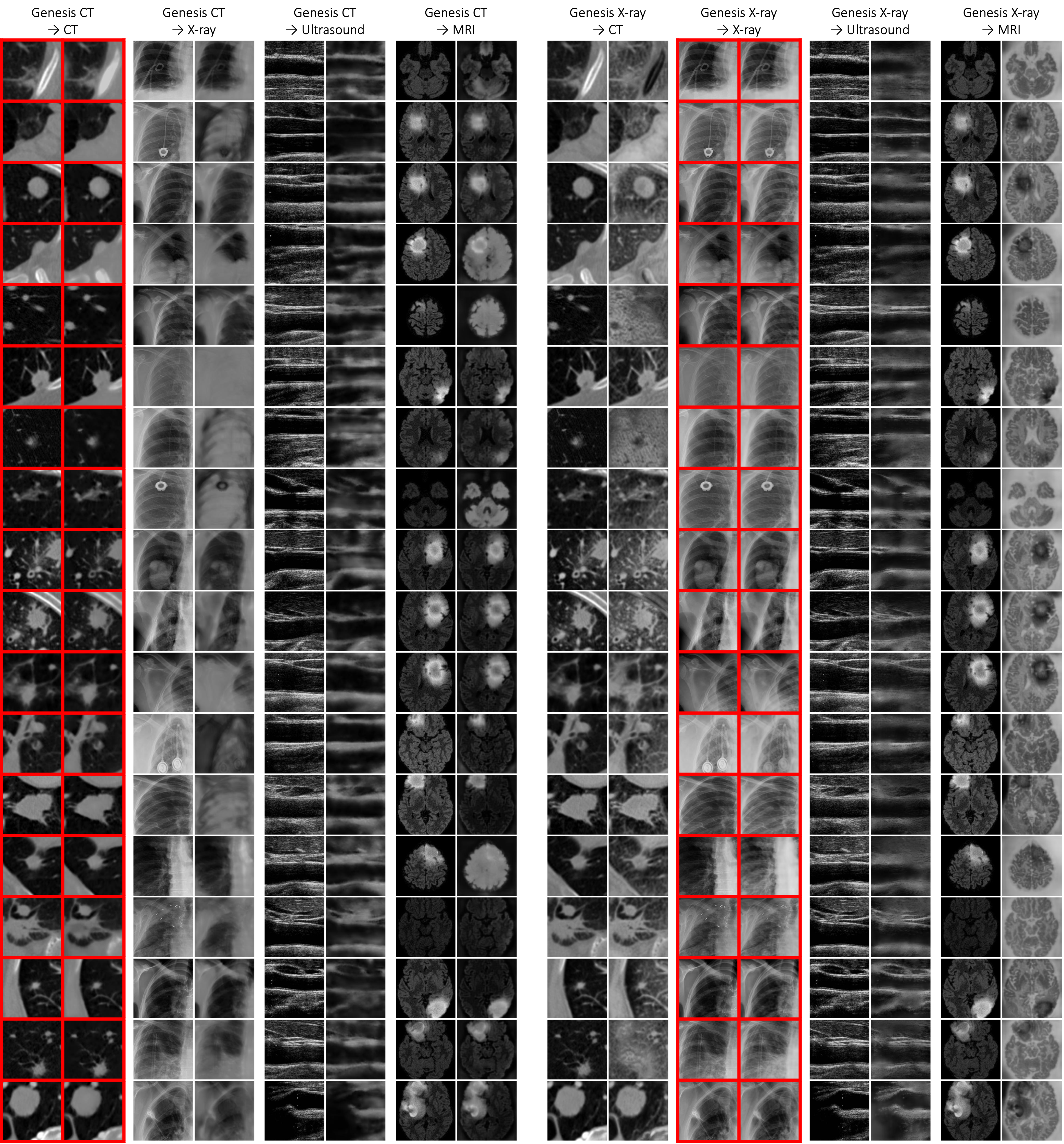}
\caption{
The left and right panels visualize the qualitative assessment of image restoration quality by Genesis CT and Genesis X-ray, respectively, across medical imaging modalities. For testing, we use the pre-trained model to directly restore images from LUNA~2016 (CT)~\citep{setio2017validation}, ChestX-ray8 (X-ray)~\citep{wang2017chestx}, CIMT (Ultrasound)~\citep{hurst10,zhou2019integrating}, and BraTS (MRI)~\citep{menze2015multimodal}. 
Though the models are only trained on single image modality, they can largely maintain the texture and structures during restoration not only within the same modality (framed in red), but also across different modalities. 
}
\label{fig:xray_across_restoration}
\end{figure*}

Since there is no such metric to directly determine the power of image representation, rather than constrain the representation, our paper aims to design an image restoration task to let the model learn generic image representation from 3D medical images. In doing so, as seen in Sec.~\ref{sec:transformation_augmentation}, we have modified the definition of a good representation.
As presented in Sec.~\ref{sec:experiments_pretraining_models_genesis}, Genesis CT and Genesis X-ray are pre-trained on LUNA~2016~\citep{setio2017validation} and ChestX-ray8~\citep{wang2017chestx}, respectively, using a series of self-supervised learning schemes with different image transformations. In this section, we have (1) illustrated more examples of our four individual transformations (\ie non-linear, local-shuffling, outer-cutout, and inner-cutout) in \figurename~\ref{fig:transformation_appendix}; (2) evaluated the power of the pre-trained model by assessing restoration quality on previously \textit{unseen} patients' images not only from the LUNA~2016 dataset (see~\figurename~\ref{fig:ct_restoration_individual_combined}), but also from different modalities, covering CT, X-ray, and MRI (see~\figurename~\ref{fig:xray_across_restoration}). The qualitative assessment shows that our pre-trained model is not merely overfitting on anatomical patterns in specific patients, but indeed can be robustly used for restoring images, thus can be generalized to many target tasks. 

To assess the image restoration quality at the time of inference, we pass the transformed images to the models that have been trained with different self-supervised learning schemes, including four individual and one combined schemes. In our visualization, the input images have undergone four individual transformations as well as eight different combined transformations, including the identity mapping (\ie no transformation). As shown in \figurename~\ref{fig:ct_restoration_individual_combined}, the combined scheme can restore the unseen image by handling a variety of transformations (framed in red), whereas the models trained with the individual scheme can only restore unseen images that have undergone the same transformation that they were trained on (framed in green). This qualitative observation is consistent with our experimental finding in Sec.~\ref{sec:individual_combination}: the combined learning scheme achieves superior and more robust results over the individual scheme in transfer learning.

In \figurename~\ref{fig:xray_across_restoration}, we have further provided a qualitative assessment of image restoration quality by Genesis CT and Genesis X-ray, across medical imaging modalities. In our visualization, the input images are selected from four different medical modalities, covering X-ray, CT, Ultrasound, and MRI. It is clear from the figure that even though the models are only trained on single image modality, they can largely maintain the texture and structures during restoration, not only within the same modality but also across different ones. These observations suggest that Models Genesis are of great potential in transferring learned image representation across diseases, organs, datasets, and modalities.

\end{document}